\documentclass[10pt,twocolumn,letterpaper]{article}

\usepackage{iccv}
\usepackage{times}
\usepackage{epsfig}
\usepackage{graphicx}
\usepackage{amsmath}
\usepackage{amssymb}

\usepackage{savesym}
\usepackage{booktabs}
\usepackage{multirow}
\savesymbol{Cross}
\usepackage{bbding}
\restoresymbol{bb}{Cross}
\usepackage{subfigure}
\usepackage{bbding}
\usepackage{lipsum}
\usepackage{xcolor}
\usepackage{colortbl}
\usepackage{marvosym}

\newcommand\blfootnote[1]{%
  \begingroup
  \renewcommand\thefootnote{}\footnote{#1}%
  \addtocounter{footnote}{-1}%
  \endgroup
}

\usepackage[pagebackref=true,breaklinks=true,letterpaper=true,colorlinks,bookmarks=false]{hyperref}

\iccvfinalcopy 

\def\iccvPaperID{****} 

\ificcvfinal\fi

\begin{document}

\definecolor{gray}{RGB}{123,123,123}
\definecolor{newborn}{RGB}{197,90,17}
\definecolor{ddetr}{RGB}{255,240,245}

\title{MeMOTR: Long-Term Memory-Augmented Transformer \\ for Multi-Object Tracking}
\author{Ruopeng Gao\textsuperscript{1} \qquad \qquad Limin Wang\textsuperscript{1,2,\Letter} \\
    $^1$State Key Laboratory for Novel Software Technology, Nanjing University \quad $^2$Shanghai AI Lab 
}

\maketitle
\ificcvfinal\thispagestyle{empty}\fi

\blfootnote{\Letter~: Corresponding author (lmwang@nju.edu.cn).}

\begin{abstract}
    As a video task, Multi-Object Tracking (MOT) is expected to capture temporal information of targets effectively.
    Unfortunately, most existing methods only explicitly exploit the object features between adjacent frames, while lacking the capacity to model long-term temporal information. 
    In this paper, we propose MeMOTR, a long-term memory-augmented Transformer for multi-object tracking.
    Our method is able to make the same object's track embedding more stable and distinguishable by leveraging long-term memory injection with a customized memory-attention layer. 
    This significantly improves the target association ability of our model. 
    Experimental results on DanceTrack show that MeMOTR impressively surpasses the state-of-the-art method by 7.9\% and 13.0\% on HOTA and AssA metrics, respectively. Furthermore, our model also outperforms other Transformer-based methods on association performance on MOT17 and generalizes well on BDD100K. Code is available at \href{https://github.com/MCG-NJU/MeMOTR}{https://github.com/MCG-NJU/MeMOTR}.
\end{abstract}

\section{Introduction}

Multi-Object Tracking (MOT)~\cite{sportsmot,MOT16,DanceTrack} aims to detect multiple objects and maintain their identities in a video stream. MOT can be applied to numerous downstream tasks, such as action recognition~\cite{Unified-MOT-Action}, behavior analysis~\cite{Behavior}, and so on. It is also an important technique for real-world applications, \eg, autonomous driving and surveillance.

According to the definition of MOT, this task can be formally divided into two parts: object detection and association.
For a long time, pedestrian tracking datasets (like MOT17~\cite{MOT16}) have had mainstream domination in the community. However, these datasets have insufficient challenges in target association because of their almost linear motion pattern.
Therefore, tracking-by-detection methods~\cite{OC-SORT, JDE, ByteTrack} achieve the state-of-the-art performance of MOT for several years. They first adopt a robust object detector (\eg, YOLOX~\cite{YOLOX}) to independently localize the objects in each frame and associate them with IoU~\cite{SORT, C-BIoU} or ReID features~\cite{ReID}. 
However, associating targets becomes a critical challenge in some complex scenarios, like group dancers~\cite{DanceTrack} and sports players~\cite{sportsmot, SoccerNet}. These similar appearances and erratic movements may cause existing methods to fail.
Recently, Transformer-based tracking methods~\cite{TrackFormer, MOTR} have introduced a new fully-end-to-end MOT paradigm. Through the interaction and progressive decoding of detect and track queries in Transformer, they simultaneously complete detection and tracking. This paradigm is expected to have greater potential for object association due to the flexibility of Transformer, especially in the above complex scenes.

Although these Transformer-based methods achieve excellent performance, they still struggle with some complicated issues, such as analogous appearances, irregular motion patterns, and long-term occlusions.
We hypothesize that more intelligent leverage of temporal information can provide the tracker a more effective and robust representation for each tracked target, thereby relieving the above issues and boosting the tracking performance.
Unfortunately, most previous methods~\cite{TrackFormer, MOTR} only exploit the image or object features between two adjacent frames, which lacking the utilization of long-term temporal information.

Based on the analysis above, in this paper, we focus on leveraging temporal information by proposing a long-term \textbf{Me}mory-augmented \textbf{M}ulti-\textbf{O}bject \textbf{T}racking method with T\textbf{R}ansformer, coined as \textbf{MeMOTR}.
We exploit detect and track embeddings to localize newborn and tracked objects via a Transformer Decoder, respectively.
Our model maintains a long-term memory with the exponential recursion update algorithm~\cite{SOT1999} for each tracked object. Afterward, we inject this memory into the track embedding, reducing its abrupt changes and thus improving the model association ability. As multiple tracked targets exist in a video stream, we apply a memory-attention layer to produce a more distinguishable representation. 
Besides, we present an adaptive aggregation to fuse the object feature from two adjacent frames to improve tracking robustness.

In addition, we argue that the learnable detection query in DETR~\cite{DETR} has no semantic information about specific objects. However, the track query in Transformer-based MOT methods like MOTR~\cite{MOTR} carries information about a tracked object. This difference will cause a semantic information gap and thus degrade the final tracking performance. Therefore, to overcome this issue, we use a light decoder to perform preliminary object detection, which outputs the detect embedding with specific semantics. Then we jointly input detect and track embeddings into the subsequent decoder to make MeMOTR tracking results more precise.

We mainly evaluate our method on the DanceTrack dataset~\cite{DanceTrack} because of its serious association challenge. Experimental results show that our method achieves the state-of-the-art performance on this challenging DanceTrack dataset, especially on association metrics (\eg, AssA, IDF1). We also evaluate our model on the traditional pedestrian tracking dataset of  MOT17~\cite{MOT16} and the multi-categories tracking dataset of BDD100K~\cite{BDD100K}.
In addition, we perform extensive ablation studies further demonstrate the effectiveness of our designs.

\begin{figure*}[t]
    \begin{center}
        \includegraphics[width=0.98\linewidth]{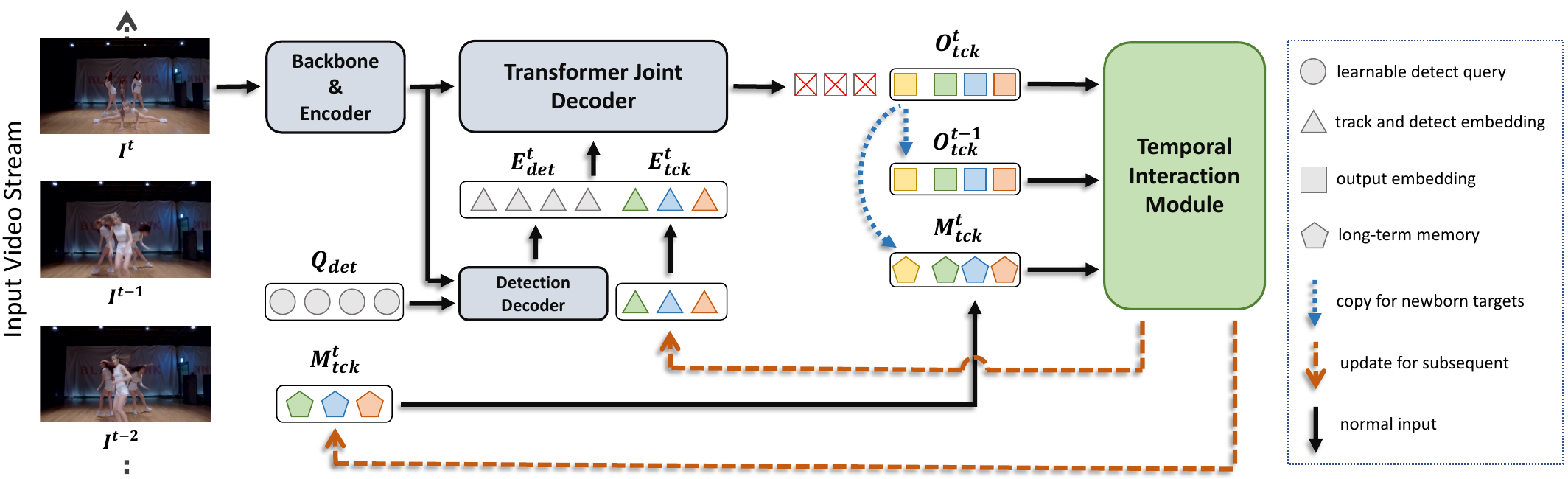}
    \end{center}
    \caption{\textbf{Overview of MeMOTR.} Like most DETR-based~\cite{DETR} methods, we exploit a ResNet-50~\cite{ResNet} backbone and a Transformer~\cite{Attention} Encoder to learn a 2D representation of an input image. We use different colors to indicate different tracked targets, and the learnable detect query $Q_{det}$ is illustrated in \textcolor{gray}{gray}. Then the Detection Decoder $\mathcal{D}_{det}$ processes the detect query to generate the detect embedding $E_{det}^t$, which aligns with the track embedding $E_{tck}^t$ from previous frames. Long-term memory is denoted as $M_{tck}^t$. The initialization process in the \textcolor{blue}{blue} dotted arrow will be applied to newborn objects. Our Long-Term Memory and Temporal Interaction Module is discussed in Section~\ref{Section:Long-Term-Memory} and~\ref{Section:TIM}. More details are illustrated in Figure~\ref{Fig:Temporal-Interaction-Module}.}
\label{Fig:Overview}
\end{figure*}

\section{Related Work}

\noindent \textbf{Tracking-by-Detection}
is a widely used MOT paradigm that has recently dominated the community. These methods always get trajectories by associating a given set of detections in a streaming video.

The objects in classic pedestrian tracking scenarios~\cite{MOT20, MOT16} always have different appearances and regular motion patterns. Therefore, appearance matching and linear motion estimation are widely used to match targets in consecutive frames.
SORT~\cite{SORT} uses the Intersection-over-Union (IoU) to match predictions of the Kalman filter~\cite{KF} and detected boxes.
Deep-SORT~\cite{Deep-SORT} applies an additional network to extract target features, then utilizes cosine distances for matching besides motion consideration in SORT~\cite{SORT}.
JDE~\cite{JDE}, FairMOT~\cite{FairMOT}, and Unicorn~\cite{Unicorn} further explore the architecture of appearance embedding and matching. ByteTrack~\cite{ByteTrack} employs a robust detector based on YOLOX~\cite{YOLOX} and reuses low-confidence detections to enhance the association ability. Furthermore, OC-SORT~\cite{OC-SORT} improves SORT~\cite{SORT} by rehabilitating lost targets.
In recent years, as a trendy framework in vision tasks, some studies~\cite{TrackByClip, GTR} have also applied Transformers to match detection bounding boxes.
Moreover, Dendorfer \etal~\cite{QuoVadis} attempt to model pedestrian trajectories by leveraging more complex motion estimation methods (like S-GAN~\cite{S-GAN}) from the trajectory prediction task.

The methods described above have powerful detection capabilities due to their robust detectors. However, although such methods have achieved outstanding performance in pedestrian tracking datasets, they are mediocre at dealing with more complex scenarios having irregular movements.
These unforeseeable motion patterns will cause the trajectory estimation and prediction module to fail.

\noindent \textbf{Tracking-by-Query} usually does not require additional post-processing to associate detection results. Unlike the tracking-by-detection paradigm mentioned above, tracking-by-query methods apply the track query to decode the location of tracked objects progressively.

Inspired by DETR-family~\cite{DETR}, most of these methods~\cite{TrackFormer, MOTR} leverage the learnable object query to perform newborn object detection, while the track query localizes the position of tracked objects.
TransTrack~\cite{TransTrack} builds a siamese network for detection and tracking, then applies an IoU matching to produce newborn targets.
TrackFormer~\cite{TrackFormer} utilizes the same Transformer decoder for both detection and tracking, then employs a non-maximum suppression (NMS) with a high IoU threshold to remove strongly overlapping duplicate bounding boxes. 
MOTR~\cite{MOTR} builds an elegant and fully end-to-end Transformer for multi-object tracking. This paradigm performs excellently in dealing with irregular movements due to the flexibility of query-based design. 
Furthermore, MQT~\cite{MQT} employs different queries to represent one tracked object and cares more about class-agnostic tracking.

However, current query-based methods typically exploit the information of adjacent frames (query~\cite{MOTR} or feature~\cite{TrackFormer} fusion). Although the track query can be continuously updated over time, most methods still do not explicitly exploit longer temporal information.
Cai \etal~\cite{MeMOT} explore a large memory bank to benefit from time-related knowledge but suffer enormous storage costs. In order to use long-term information, we propose a long-term memory to stabilize the tracked object feature over time and a memory-attention layer for a more distinguishable representation. 
Our experiments further approve that this approach significantly improves association performance in MOT.

\section{Method}

\subsection{Overview} \label{Section:Overview}
We propose the \textbf{MeMOTR}, a long-term memory-augmented Transformer for multi-object tracking.
Different from most existing methods~\cite{TrackFormer, MOTR} that only explicitly utilize the states of tracked objects between adjacent frames, our core contribution is to build a \textit{long-term memory} (in Section~\ref{Section:Long-Term-Memory}) that maintains the long-term temporal feature for each tracked target, together with a \textit{temporal interaction module (TIM)} that effectively injects the temporal information into subsequent tracking processes.

Like most DETR-family methods~\cite{DETR}, we use a ResNet-50~\cite{ResNet} backbone and a Transformer Encoder to produce the image feature of an input frame $I^t$. As shown in Figure~\ref{Fig:Overview}, the learnable detect query $Q_{det}$ is fed into the \textit{Detection Decoder} $\mathcal{D}_{det}$ (in Section~\ref{Section:Detection-Decoder}) to generate the detect embedding $E_{det}^t$ for the current frame. Afterward, by querying the encoded image feature with $[E_{det}^t, E_{tck}^t]$, the Transformer Joint Decoder $\mathcal{D}_{joint}$ produces the corresponding output $[\hat{O}_{det}^t, \hat{O}_{tck}^t]$. For simplicity, we merge the newborn objects in $\hat{O}_{det}^t$ (yellow box) with tracked objects' output $\hat{O}_{tck}^t$, denoted by $O_{tck}^t$. 
Afterward, we predict the classification confidence $c_i^t$ and bounding box $b_i^t$ corresponding to the $i^{th}$ target from the output embeddings.
Finally, we feed the output from adjacent frames $[O_{tck}^t, O_{tck}^{t-1}]$ and the long-term memory $M_{tck}^t$ into the Temporal Interaction Module, updating the subsequent track embedding $E_{tck}^{t+1}$ and long-term memory $M_{tck}^{t+1}$. The details of our components will be elaborated in the following sections.

\subsection{Detection Decoder} \label{Section:Detection-Decoder}
In the previous Transformer-based methods~\cite{TrackFormer, MOTR}, the learnable detect query and the previous track query are jointly input to Transformer Decoder from scratch. This simple idea extends the end-to-end detection Transformer~\cite{DETR} to multi-object tracking. Nonetheless, we argue that this design may cause misalignment between detect and track queries.
As discussed in numerous works~\cite{DETR, DAB-DETR}, the learnable object query in DETR-family plays a role similar to a learnable anchor with little semantic information. On the other hand, track queries have specific semantic knowledge to resolve their category and bounding boxes since they are generated from the output of previous frames.

Therefore, as illustrated in Figure~\ref{Fig:Overview}, we split the original Transformer Decoder into two parts. The first decoder layer is used for detection, and the remaining five layers are used for joint detection and tracking. These two decoders have the same structure but different inputs.
The Detection Decoder $\mathcal{D}_{det}$ takes the original learnable detect query $Q_{det}$ as input and generates the corresponding detect embedding $E_{det}^t$, carrying enough semantic information to locate and classify the target roughly.
After that, we concatenate the detect and track embedding together and feed them into the Joint Decoder $\mathcal{D}_{joint}$.

\subsection{Long-Term Memory} \label{Section:Long-Term-Memory}
Unlike previous methods~\cite{MQT, MOTR} that only exploit adjacent frames' information, we explicitly introduce a \textit{long-term memory} $M_{tck}^t$ to maintain longer temporal information for tracked targets.
When a newborn object is detected, we initialize its long-term memory with the current output.

It should be noted that in a video stream, objects only have minor deformation and movement in consecutive frames. Thus, we suppose the semantic feature of a tracked object changes only slightly in a short time. In the same way, our long-term memory should also update smoothly over time. Inspired by~\cite{SOT1999}, we apply a simple but effective running average with exponentially decaying weights to update long-term memory $M_{tck}^t$:

\begin{equation}
    \widetilde{M}_{tck}^{t+1} = (1 - \lambda) M_{tck}^t + \lambda \cdot O_{tck}^t,
    \label{Eq:Memory-Update}
\end{equation}
where $\widetilde{M}_{tck}^{t+1}$ is the new long-term memory for the next frame. The memory update rate $\lambda$ is experimentally set to $0.01$, following the assumption that the memory changes smoothly and consistently in consecutive frames. We also tried some other values in Table~\ref{Table:Long-Memory-Lambda}.

\subsection{Temporal Interaction Module} \label{Section:TIM}

\begin{figure}[t]
    \begin{center}
        \includegraphics[width=0.7\linewidth]{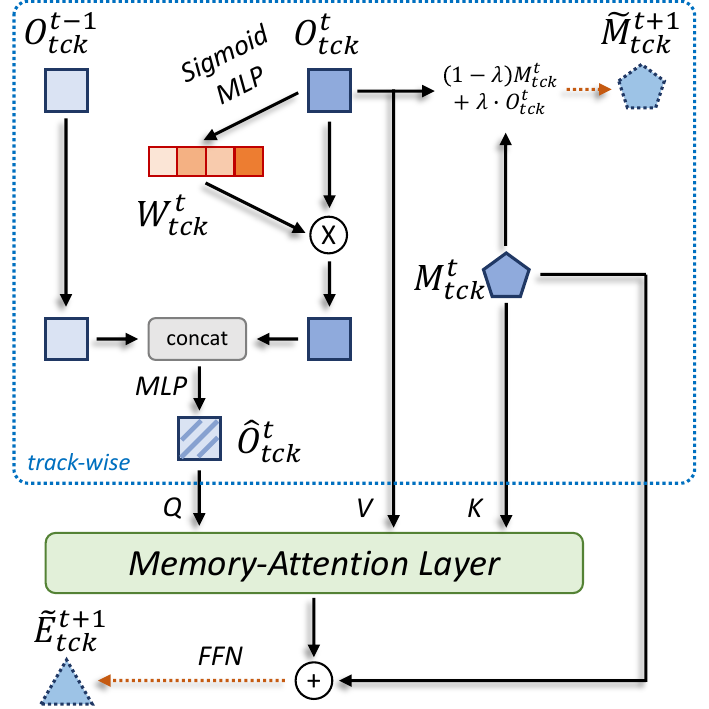}
    \end{center}
    \caption{Illustration of \textbf{Temporal Interaction Module.} $\widetilde{E}_{tck}^{t+1}$ and $\widetilde{M}_{tck}^{t+1}$ are the prediction of $E_{tck}^t$ and $M_{tck}^t$ for the next frame, respectively.}
\label{Fig:Temporal-Interaction-Module}
\end{figure}

\noindent \textbf{Adaptive Aggregation for Temporal Enhancement.}
Issues such as blurring or occlusion are often seen in a video stream. An intuitive idea to solve this problem is using multi-frame features to enhance the single-frame representation. Therefore, we fuse the outputs from two adjacent frames with an adaptive aggregation algorithm in our MeMOTR. Due to occlusions and blurring, the output embedding $O_{tck}^t$ of the current frame may be unreliable. Thus, as illustrated in Figure~\ref{Fig:Temporal-Interaction-Module}, we generate a \textit{channel-wise weight} $W_{tck}^t$ for each tracked instance to alleviate this problem:

\begin{equation}
    W_{tck}^t = \mathrm{Sigmoid}(\mathrm{MLP}(O_{tck}^t)).
    \label{Eq:Adaptive-Weight}
\end{equation}

We multiply this weight $W_{tck}^t$ with the current output $O_{tck}^t$ and then concatenate the result with $O_{tck}^{t-1}$ from the previous frame. Furthermore, we apply a two-layer MLP to produce the fusion outcome $\widetilde{O}_{tck}^t$. This adaptive aggregation enhances the target representation with short-term temporal modeling. 

However, we do not use the above channel-wise weight for previous output $O_{tck}^{t-1}$. As we will discuss in Section~\ref{Section:Inference}, there is a difference between $O_{tck}^t$ and $O_{tck}^{t-1}$. During inference, we employ a score threshold $\tau_{tck}$ to guarantee that $O_{tck}^{t-1}$ is relatively reliable. 
Therefore, we input it entirely into the subsequent fusion step without the adaptive weight.

\vspace{2mm}
\noindent \textbf{Generate Track Embedding.}
As discussed in Section~\ref{Section:Overview}, we exploit the track embedding $E_{tck}^t$ to produce the location and category of each tracked target. Therefore, generating more reliable and distinguishable track embedding is the key to improving tracking performance. Our processing is illustrated in Figure~\ref{Fig:Temporal-Interaction-Module}.

Since there are multiple similar objects in the same frame, we believe that learning more discriminative representations is also crucial to the tracker. Thus we employ a Multi-Head Attention~\cite{Attention} structure called \textit{memory-attention layer} to achieve this interaction between different trajectories. Due to the reliability of long-term memory $M_{tck}^t$, we use it as $K$, and the aggregation $\hat{O}_{tck}^t$ and the output embedding $O_{tck}^t$ as $Q$ and $V$, respectively.

After that, we combine long-term memory $M_{tck}^t$ and the result of memory-attention layer using addition, then input into an FFN network to predict the subsequent track embedding $\widetilde{E}_{tck}^{t+1}$. As shown in Equation~(\ref{Eq:Memory-Update}), long-term memory is gradually changing over time. Therefore, by incorporating information from long-term memory, the track embedding $\widetilde{E}_{tck}^{t+1}$ avoids abrupt changes that may cause association mistakes in consecutive frames. This design significantly improves the performance of object association, corroborated by ablation experiments shown in Table~\ref{Table:Temporal-Interaction}.

\subsection{Inference Details} \label{Section:Inference}

At time step $t$, we jointly input the learnable detect query $Q_{det}$ and track embedding $E_{tck}^t$ $(E_{tck}^0 = \emptyset)$ into our model to produce detection and tracking results, respectively. The detection result with a confidence score of more than $\tau_{det}$ will transform into a newborn object.

Target occlusion is a common issue in multi-object tracking task. 
If a tracked object is lost (confidence $\leq \tau_{tck}$) in the current frame, we do not directly remove its track embedding but mark it as $inactive$ trajectory.
Afterward, the inactive target will be removed entirely after $\mathcal{T}_{miss}$ frames.

It is worth noting that we do not update the track embedding and long-term memory at every time step for each object. Instead, we choose to update those track embedding with high confidence. The choice of update threshold $\tau_{next}$ yields the following formulation for updating:

\begin{equation}
    [E^{t+1}_i, M^{t+1}_i] = 
    \begin{cases}
        [\widetilde{E}^{t+1}_i, \widetilde{M}^{t+1}_i], & c_i^t > \tau_{next} \\
        [E^{t}_i, M^{t}_i], & c_i^t \leq \tau_{next}
    \end{cases},
\end{equation}
where $i$ is the target index, $t$ is the frame index, and $c_i^t$ is the predicted classification confidence of the $i^{th}$ object at time step $t$. $\widetilde{E}^{t+1}_i$ and $\widetilde{M}^{t+1}_i$ are the predictions of track embedding and long-term memory, generated by Temporal Interaction Module shown in Figure~\ref{Fig:Temporal-Interaction-Module}. For simplicity, we set $\tau_{det} = \tau_{tck} = \tau_{next} = 0.5$ in our experiments. $\mathcal{T}_{miss}$ is set to $30$, $15$ and $10$ on DanceTrack, MOT17 and BDD100K, respectively.

\section{Experiments}

\subsection{Datasets and Metrics}

\noindent \textbf{Datasets.}
We mainly evaluate MeMOTR on the DanceTrack~\cite{DanceTrack} dataset since they have more severe association challenges than traditional pedestrian tracking datasets. For a comprehensive evaluation, we also conduct experiments on MOT17~\cite{MOT16} and BDD100K~\cite{BDD100K}.

\noindent \textbf{Metrics.}
Because of providing a balanced way to measure both detection and association performance explicitly, we use Higher Order Metric for Evaluating Multi-Object Tracking (HOTA)~\cite{HOTA} to evaluate our method, especially analyzing our memory mechanism using Association Accuracy (AssA).
We also list the MOTA~\cite{MOTA} and IDF1~\cite{IDF1} metrics in our experimental results.

\subsection{Implementation Details} \label{Section:Details}
Following the settings in MOTR~\cite{MOTR}, we use several data augmentation methods, such as random resize and random crop. The shorter side of the input image is resized to 800, and the maximum size is restricted to 1536. 

We built MeMOTR upon DAB-Deformable-DETR~\cite{DAB-DETR} with a ResNet50~\cite{ResNet} backbone and initialize our model with the official DAB-Deformable-DETR~\cite{DAB-DETR} weights pre-trained on the COCO~\cite{COCO} dataset. 
We suggest that the anchor-based position-prior from DAB-Deformable-DETR is quite effective due to the tracked box's smoothness in time and can be further exploited in future works.
We also provide the results of our model based on Deformable-DETR~\cite{DeformableDETR} for fair comparison in Table~\ref{Table:DanceTrack-SOTA}.
Our models are conducted on PyTorch with 8 NVIDIA Tesla V100 GPUs. 
By using PyTorch gradient checkpoint technology, we implement a memory-optimized version that can also be trained on NVIDIA GPUs with less than 10GB GPU memory.
The batch size is set to 1 per GPU, and each batch contains a video clip with multiple frames. Within each clip, video frames are sampled with random intervals from 1 to 10.
We use the AdamW optimizer with the initial learning rate of $2.0 \times 10^{-4}$. 
During training, we filter out the tracked target lower than the score threshold $\tau_{update}=0.5$ and IoU threshold $\tau_{IoU}=0.5$. 

On \textbf{DanceTrack}~\cite{DanceTrack}, we train MeMOTR for $18$ epochs on the train set and drop the learning rate by a factor of $10$ at the $12^{th}$ epoch. Firstly, we use two frames within a clip for training. And then increase the clip frames to 3, 4, and 5 at the $6^{th}$, $10^{th}$, and $14^{th}$ epochs, respectively.
On \textbf{MOT17}~\cite{MOT16}, due to the small train set (about 5K frames), it is easy to cause overfitting problems. Therefore, we add CrowdHuman~\cite{CrowdHuman} validation set to build a joint train set with MOT17 training data. CrowdHuman val set provides about 4K static images. Therefore, we apply random shifts from CenterTrack~\cite{CenterTrack} to generate pseudo trajectories. Finally, we train MeMOTR for $130$ epochs, and the learning rate decays by a factor of $10$ at the $120^{th}$ epoch. The initial length of the training video clip is $2$ and gradually increases to $3$ and $4$ at the $60^{th}$ and $100^{th}$ epochs, respectively.
On \textbf{BDD100K}~\cite{BDD100K}, we modify the sampling length at the $6^{th}$ and $10^{th}$ epochs and totally train 14 epochs while reducing the learning rate at the $12^{th}$ epoch.

\begin{table}[t] \small
  \begin{center}
  \setlength{\tabcolsep}{1.2mm}{
    \begin{tabular}{l|ccccc}
      \toprule[2pt]
      Methods & HOTA & DetA & AssA & MOTA & IDF1 \\
      \midrule[1pt]
      \textit{w/o extra data:} \\
      FairMOT~\cite{FairMOT} & 39.7 & 66.7 & 23.8 & 82.2 & 40.8 \\
      CenterTrack~\cite{CenterTrack} & 41.8 & 78.1 & 22.6 & 86.8 & 35.7 \\
      TraDeS~\cite{TraDeS} & 43.3 & 74.5 & 25.4 & 86.2 & 41.2 \\
      TransTrack~\cite{TransTrack} & 45.5 & 75.9 & 27.5 & 88.4 & 45.2 \\
      ByteTrack~\cite{ByteTrack} & 47.7 & 71.0 & 32.1 & 89.6 & 53.9 \\
      GTR~\cite{GTR} & 48.0 & 72.5 & 31.9 & 84.7 & 50.3 \\
      QDTrack~\cite{QDTrack} & 54.2 & 80.1 & 36.8 & 87.7 & 50.4 \\
      MOTR~\cite{MOTR} & 54.2 & 73.5 & 40.2 & 79.7 & 51.5 \\
      OC-SORT~\cite{OC-SORT} & 55.1 & 80.3 & 38.3 & \bf 92.0 & 54.6 \\
      C-BIoU~\cite{C-BIoU} & 60.6 & \bf 81.3 & 45.4 & 91.6 & 61.6 \\
      MeMOTR\textsuperscript{*} (ours) & 63.4 & 77.0 & 52.3 & 85.4 & 65.5 \\
      MeMOTR (ours) & \bf 68.5 & 80.5 & \bf 58.4 & 89.9 & \bf 71.2 \\
      \midrule[1pt]
      \textit{\textcolor{gray}{with extra data:}} \\
      \textcolor{gray}{MT\underline{~~}IoT}~\cite{MT_IoT} & \textcolor{gray}{66.7} & \textcolor{gray}{84.1} & \textcolor{gray}{53.0} & \textcolor{gray}{94.0} & \textcolor{gray}{70.6} \\
      \textcolor{gray}{MOTRv2}~\cite{MOTRv2} & \textcolor{gray}{69.9} & \textcolor{gray}{83.0} & \textcolor{gray}{59.0} & \textcolor{gray}{91.9} & \textcolor{gray}{71.7} \\
      \bottomrule[2pt]
    \end{tabular}
  }
  \end{center}
  \caption{Performance comparison with state-of-the-art methods on the DanceTrack~\cite{DanceTrack} test set. Results for existing methods are from DanceTrack~\cite{DanceTrack}. MeMOTR\textsuperscript{*} means the result based on standard Deformable-DETR.}
  \label{Table:DanceTrack-SOTA}
\end{table}

\subsection{Comparison on the DanceTrack Dataset}

Since DanceTrack~\cite{DanceTrack} is a dataset with various motions that cannot be modeled by classic linear motion estimation~\cite{SORT, OC-SORT}, it provides a better choice to verify our tracking performance, especially the association performance.

We compare MeMOTR with the state-of-the-art methods on the DanceTrack~\cite{DanceTrack} test set in Table~\ref{Table:DanceTrack-SOTA}. Our method achieves $68.5$ HOTA and gains a vast lead on the AssA metric ($58.4$ AssA), even surpassing some methods~\cite{MT_IoT} that use additional datasets for training. 
Due to the limitations of the linear motion estimation module, some tracking-by-detection methods, for example, ByteTrack~\cite{ByteTrack}, although they can achieve great detection results ($71.0$ DetA), still cannot handle complex object association problems ($32.1$ AssA). However, their MOTA metrics are still high because MOTA overemphasizes detection performance.

Our temporal interaction module, shown in Figure~\ref{Fig:Temporal-Interaction-Module}, leverages temporal information gracefully and efficiently.
Moreover, the separated detection decoder $\mathcal{D}_{det}$ discussed in Section~\ref{Section:Detection-Decoder} alleviates the conflicts between detection and tracking tasks. Therefore, we earn an impressive association performance ($58.4$ AssA and $71.2$ IDF1) and competitive detection performance ($80.5$ DetA) compared with the state-of-the-art methods.
We further prove our components' effectiveness in Section~\ref{Section:Ablation}.

\begin{table}[t] \small
  \begin{center}
  \setlength{\tabcolsep}{1.0mm}{
    \begin{tabular}{l|ccccc}
      \toprule[2pt]
      Methods & HOTA & DetA & AssA & MOTA & IDF1 \\
      \midrule[1pt]
      \textit{CNN based:} \\
          Tracktor++~\cite{Tracktor++} & 44.8 & 44.9 & 45.1 & 53.5 & 52.3 \\
          CenterTrack~\cite{CenterTrack} & 52.2 & 53.8 & 51.0 & 67.8 & 64.7 \\
          TraDeS~\cite{TraDeS} & 52.7 & 55.2 & 50.8 & 69.1 & 63.9 \\
          QDTrack~\cite{QDTrack} & 53.9 & 55.6 & 52.7 & 68.7 & 66.3 \\
          GTR~\cite{GTR} & 59.1 & 61.6 & 57.0 & 75.3 & 71.5 \\
          FairMOT~\cite{FairMOT} & 59.3 & 60.9 & 58.0 & 73.7 & 72.3 \\
          DeepSORT~\cite{Deep-SORT} & 61.2 & 63.1 & 59.7 & 78.0 & 74.5 \\
          SORT~\cite{SORT} & 63.0 & 64.2 & 62.2 & 80.1 & 78.2 \\
          ByteTrack~\cite{ByteTrack} & 63.1 & 64.5 & 62.0 & 80.3 & 77.3 \\
          Quo Vadis~\cite{QuoVadis} & 63.1 & 64.6 & 62.1 & 80.3 & 77.7 \\
          OC-SORT~\cite{OC-SORT} & 63.2 & 63.2 & 63.4 & 78.0 & 77.5 \\
          C-BIoU~\cite{C-BIoU} & 64.1 & 64.8 & 63.7 & 81.1 & 79.7 \\
      \midrule[1pt]
      \textit{Transformer based:} \\
          TrackFormer~\cite{TrackFormer} & / & / & / & 74.1 & 68.0 \\
          TransTrack~\cite{TransTrack} & 54.1 & \bf 61.6 & 47.9 & \bf 74.5 & 63.9 \\
          TransCenter~\cite{TransCenter} & 54.5 & 60.1 & 49.7 & 73.2 & 62.2 \\
          MeMOT~\cite{MeMOT} & 56.9 & / & 55.2 & 72.5 & 69.0 \\
          MOTR~\cite{MOTR} & 57.2 & 58.9 & 55.8 & 71.9 & 68.4 \\
          MeMOTR (ours) & \textbf{58.8} & 59.6 & \textbf{58.4} & 72.8 & \textbf{71.5} \\
      \midrule[1pt]
      \textit{\textcolor{gray}{Hybird based:}} \\
          \textcolor{gray}{MOTRv2}~\cite{MOTRv2} & \textcolor{gray}{62.0} & \textcolor{gray}{63.8} & \textcolor{gray}{60.6} & \textcolor{gray}{78.6} & \textcolor{gray}{75.0} \\
      \bottomrule[2pt]
    \end{tabular}
  }
  \end{center}
  \caption{Performance comparison with state-of-the-art methods on the MOT17~\cite{MOT16} test set. The best performance among the Transformer-based methods is marked in \textbf{bold}. MOTRv2~\cite{MOTRv2} is marked in \textcolor{gray}{hybrid} since their YOLOX~\cite{YOLOX} proposals.}
  \label{Table:MOT17-SOTA}
\end{table}

\subsection{Comparison on the MOT17 Dataset}

In order to make a comprehensive comparison, we also evaluate our method on the classic pedestrian tracking benchmark. Table~\ref{Table:MOT17-SOTA} compares our method with state-of-the-art methods on the MOT17~\cite{MOT16} test set.

Recent tracking-by-detection methods~\cite{C-BIoU, ByteTrack} exploit robust detectors (like YOLOX~\cite{YOLOX}) to achieve really excellent detection performance (up to $64.8$ DetA). Since performance on MOT17 overemphasizes detection performance, these methods perform immensely well. In this regard, there is still a massive gap in the detection performance of Transformer-based methods~\cite{MeMOT, MOTR} because too many dense and small object predictions are involved.
In addition, the joint query in a shared Transformer decoder produces tracking and detection simultaneously, which may cause internal conflicts. The detect query is inhibited by the track query in the self-attention~\cite{Attention} structure, limiting the ability to detect newborn objects, especially those close to tracked targets, and vice versa. Because of this, TransTrack~\cite{TransTrack} achieves significantly better detection performance ($61.6$ DetA) due to its siamese network structure. This architecture decouples tracking and detection to resolve the above conflict, but its simple post-processing matching algorithm decreases its association performance.

On the other hand, we found that Transformer-based methods~\cite{MOTR} suffer from serious overfitting problems in MOT17~\cite{MOT16} because of the tiny train set, which only contains about 5K frames. 
Although we use an additional CrowdHuman validation set for training mentioned in Section~\ref{Section:Details}, severe overfitting still happens. Thus, we get $\sim$$90.0$ HOTA and $\sim$$95.0$ MOTA on the train set. However, too much additional training data can lead to inductive bias toward static people. Therefore, we argue that the train set of MOT17 is too small to train our model completely.

Eventually, our method slightly improves the performance on the MOT17 test set to $58.8$ HOTA. We gain competitive detection accuracy (DetA) compared with other Transformer-based methods. In particular, we improved the performance of object association, which AssA and IDF1 metrics can reflect. 
As a method that also uses the memory mechanism, our MeMOTR achieves higher AssA and IDF1, surpassing MeMOT~\cite{MeMOT} by $3.2\%$ and $2.5\%$, respectively. These experimental results further validate the effectiveness of our method.

\begin{table}[t] \small
  \begin{center}
  \setlength{\tabcolsep}{0.6mm}{
    \begin{tabular}{l|ccccc}
      \toprule[2pt]
      Method & mTETA & mHOTA & mLocA & mAssocA & mAssA \\
      \midrule[1pt]
      QDTrack~\cite{QDTrack} & 47.8 & / & 45.9 & 48.5 & / \\
      DeepSORT~\cite{Deep-SORT} & 48.0 & / & 46.4 & 46.7 & / \\
      MOTR~\cite{MOTR} & 50.7 & 37.0 & 35.8 & 51.0 & 47.3 \\
      TETer~\cite{TETer} & 50.8 & / & \bf 47.2 & 52.9 & / \\
      MeMOTR (ours) & \bf 53.6 & \bf 40.4 & 38.1 & \bf 56.7 & \bf 52.6 \\
      \bottomrule[2pt]
    \end{tabular}
  }
  \end{center}
  \caption{Performance comparison on the BDD100K~\cite{BDD100K} val set. Results for existing methods are from~\cite{TETer}.}
  \label{Table:BDD100K-SOTA}
\end{table}

\begin{figure}[t]
    \begin{center}
        \includegraphics[width=0.85\linewidth]{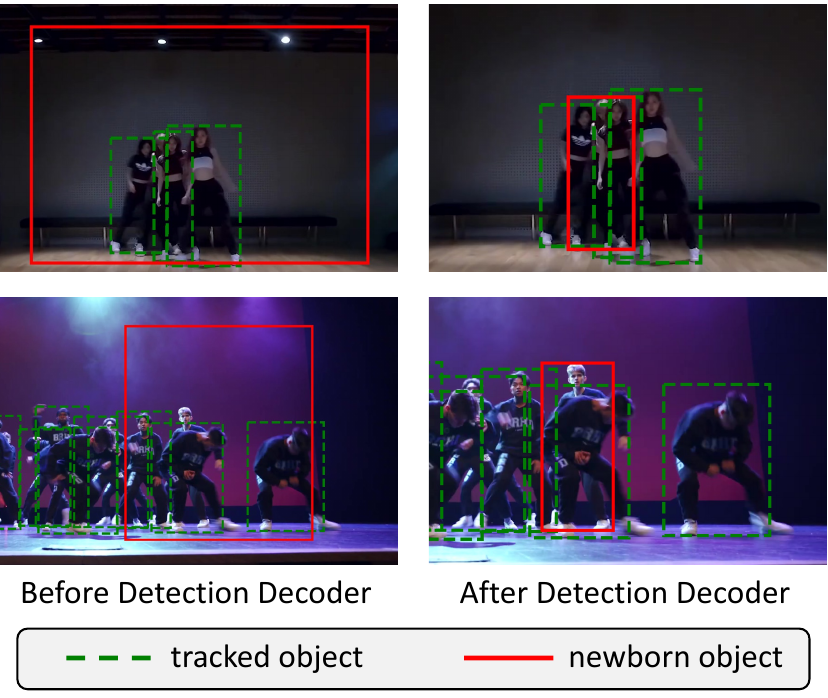}
    \end{center}
    \caption{Visualize the anchors of tracked and newborn targets before (left) and after (right) the separated detection decoder $\mathcal{D}_{det}$.}
\label{Fig:Anchors}
\end{figure}

\subsection{Comparison on the BDD100K Dataset}

In order to evaluate our method in multi-category scenarios, we also evaluate our method on the BDD100K~\cite{BDD100K} val set in Table~\ref{Table:BDD100K-SOTA}. To better assess multi-class tracking performance, we leverage Tracking-Every-Thing Accuracy (TETA) metric for ranking as they did in BDD100K MOT Challenge 2023. We re-evaluate MOTR~\cite{MOTR} on the new metrics using the official model for comparison.

The results show that our method can generalize well to multi-class scenarios and achieve impressive performance (53.6 TETA), especially in association (56.7 mAssocA), further demonstrating the effectiveness of our proposed method for associating targets. 

\subsection{Ablation Study} \label{Section:Ablation}

In this section, we study several components of our model, such as the long-term memory, adaptive aggregation, memory-attention layer, and separated detection decoder.
Since our main contribution is a better utilization of temporal information, we choose to conduct ablation experiments on DanceTrack~\cite{DanceTrack} due to its more severe object association challenges. 
On the other hand, DanceTrack~\cite{DanceTrack} has more extensive training data to avoid severe overfitting (about $10 \times$ compared with MOT17~\cite{MOT16}). We train our model on the train set and evaluate it on the official val set. 

\noindent \textbf{Detection Decoder.}
For joint tracking and detection, the tracking-by-query paradigm processes detect and track queries in a shared Transformer decoder from scratch.
However, the track query has rich semantic content from the previous tracked targets, in contrast to the object query for detection. 
On the other hand, learnable detect anchors often cover a larger range to find potential targets. In contrast, the anchor of tracked object pays more attention to a small area where the target appeared in the previous frame. 
This may cause a large gap between the anchors of tracked and newborn objects, as we visualized in Figure~\ref{Fig:Anchors} (left).

\begin{table}[t] \small
  \begin{center}
    \begin{tabular}{c|ccccc}
    \toprule[2pt]
    $\mathcal{L}_{\mathcal{D}_{det}}$ & HOTA & DetA & AssA & MOTA & IDF1 \\ 
    \midrule[1pt]
      0 & 62.1 & 74.3 & 52.2 & 83.1 & 65.6 \\ 
      1 & \bf 63.9 & \bf 74.6 & \bf 55.0 & \bf 83.4 & \bf 67.1 \\
      2 & 63.2 & 73.8 & 54.3 & 81.9 & 65.8 \\
    \bottomrule[2pt]
    \end{tabular}
  \end{center}
  \caption{Ablation experiments on the layers of the separate Detection Decoder, which is denoted as $\mathcal{L}_{\mathcal{D}_{det}}$.}
  \label{Table:Detection-Decoder}
\end{table}

\begin{table}[t] \small
  \begin{center}
    \begin{tabular}{cc|cccc}
    \toprule[2pt]
    $O_{t-1}$ & $W_{tck}^t$ & HOTA & DetA & AssA & IDF1 \\ 
    \midrule[1pt]
       &  & 62.3 & 73.9 & 52.7 & 64.6 \\ 
      & \Checkmark & 62.4 & 74.5 & 52.5 & 64.6 \\
    \midrule[1pt]
      \Checkmark &  & 62.7 & 74.4 & 53.1 & 65.3 \\
      \Checkmark & \Checkmark & \bf 63.9 & \bf 74.6 & \bf 55.0 & \bf 67.1 \\
    \bottomrule[2pt]
    \end{tabular}
  \end{center}
  \caption{Ablations on different designs of Adaptive Aggregation.}
  \label{Table:Adaptive-Aggregation}
\end{table}

In this paper, we apply a separated Transformer decoder layer to perform preliminary target detection. The output $E_{det}^t$ of this Detection Decoder $\mathcal{D}_{det}$ will be better aligned with the track embedding $E_{tck}^t$ generated by the previous frame to improve the tracking performance. We experimentally confirmed the effectiveness of this design, as shown in Table~\ref{Table:Detection-Decoder}. Using only one separate Detection Decoder layer dramatically improves HOTA and AssA metrics by $1.8\%$ and $2.8\%$, respectively. However, continuing to increase the layers of the Detection Decoder will reduce the refinement steps of track embeddings, thus slightly weakening the association performance. 
Furthermore, we visualize the bounding boxes after Detection Decoder in Figure~\ref{Fig:Anchors} (right). This indicates that $\mathcal{D}_{det}$ is able to locate objects roughly.

\noindent \textbf{Adaptive Aggregation.}
In Section~\ref{Section:TIM}, we design an adaptive aggregation, which dynamically fuses object features from adjacent frames. We ablate this structure in Table~\ref{Table:Adaptive-Aggregation}.

The first two results only use the current output $O_{tck}^t$ to generate the temporal aggregation $\hat{O}_{tck}^t$. In contrast, the next two lines fuse the previous output $\hat{O}_{tck}^{t-1}$ into $\hat{O}_{tck}^t$. 
Introducing $O_{tck}^{t-1}$ provides additional object features from neighboring frames, thus improving tracking performance. We suppose this offers a complementary feature augmentation that can combat video ambiguity and uncertainty. 

Furthermore, we explore the impact of the dynamic weight $W_{tck}^t$. As shown in Table~\ref{Table:Adaptive-Aggregation}, it only provides a little boost without $O_{tck}^{t-1}$ from the previous. We explain that dynamic $W_{tck}^t$ leads to missing information without complementary features from previous $O_{tck}^{t-1}$. 
The result of the last row shows that utilizing both dynamic weight $W_{tck}^t$ and previous output $O_{tck}^{t-1}$ produces significantly better performance, with $+2.6\%$ HOTA and $+2.3\%$ AssA.

\begin{table}[t] \small
  \begin{center}
    \begin{tabular}{cc|ccccc}
    \toprule[2pt]
        $M_{tck}^t$ & $\mathit{attn}$ & HOTA & DetA & AssA & IDF1 \\ 
    \midrule[1pt]
       \multicolumn{2}{c|}{\textit{na\"ive}} & 61.1 & 74.2 & 50.6 & 63.7 \\
    \midrule[1pt]
         &  & 61.9 & 73.8 & 52.1 & 64.1 \\
        \Checkmark & & 62.5 & 74.2 & 52.9 & 64.7 \\
    \midrule[1pt]
         & \Checkmark & 61.1 & 74.0 & 50.7 & 62.4 \\
        \Checkmark & \Checkmark & \bf 63.9 & \bf 74.6 & \bf 55.0 & \bf 67.1 \\
    \bottomrule[2pt]
    \end{tabular}
  \end{center}
  \caption{Ablation study of long-term memory $M_{tck}^t$ and memory-attention layer $\mathit{attn}$. \textit{na\"ive} means a naive baseline with a single FFN for $O_{tck}^t$ to generate the track embedding $\widetilde{E}_{tck}^{t+1}$.}
  \label{Table:Temporal-Interaction}
\end{table}

\begin{table}[t] \small
  \begin{center}
  \setlength{\tabcolsep}{1.2mm}{
    \begin{tabular}{l|ccccc}
    \toprule[2pt]
    $\lambda$ & HOTA$\uparrow$ & DetA$\uparrow$ & AssA$\uparrow$ & IDF1$\uparrow$ & IDsw $\downarrow$ \\
    \midrule[1pt]
    0.005 & 62.6 & 74.2 & 53.1 & 65.1 & 1383 \\
    0.01  & \textbf{63.9} & 74.6 & \textbf{55.0} & \textbf{67.1} & 1237 \\
    0.02  & 63.2 & 75.0 & 53.5 & 66.0 & \textbf{1213} \\
    0.04  & 63.5 & \textbf{75.1} & 54.0 & 66.3 & 1295 \\
    \bottomrule[2pt]
    \end{tabular}
  }
  \end{center}
  \caption{Ablation study on the long-term memory update ratio $\lambda$.}
  \label{Table:Long-Memory-Lambda}
\end{table}

\begin{figure*}[t]
	\centering
	    \subfigure[\textit{w/o memory} and \textit{w/o attention}]{\includegraphics[width=0.24\linewidth]{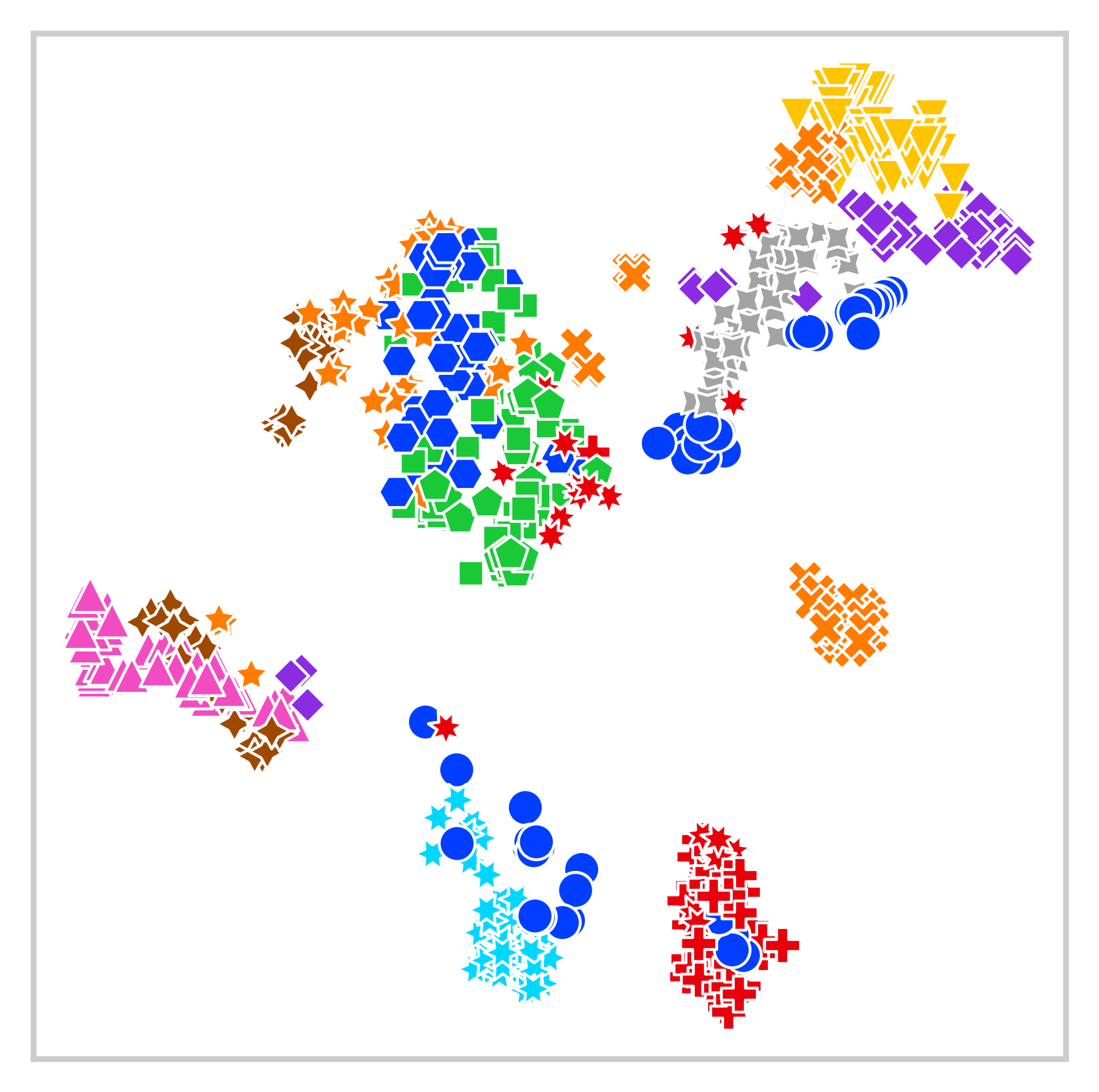} \label{Fig:Track-Embedding:nothing}} 
        \subfigure[\textit{w/o memory} and \textit{w/ attention}]{\includegraphics[width=0.24\linewidth]{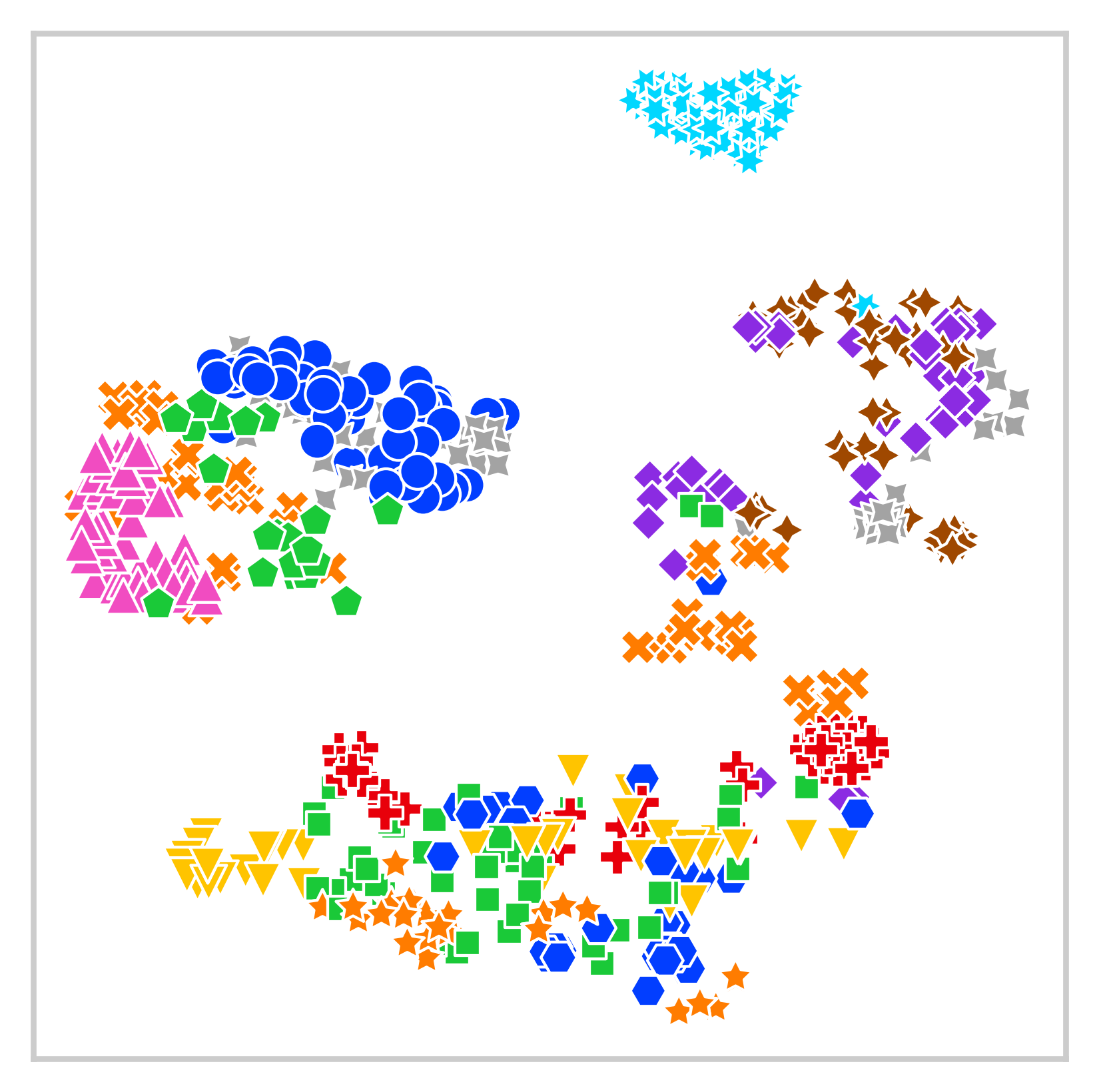} \label{Fig:Track-Embedding:attention}}
        \subfigure[\textit{w/ memory} and \textit{w/o attention}]{\includegraphics[width=0.24\linewidth]{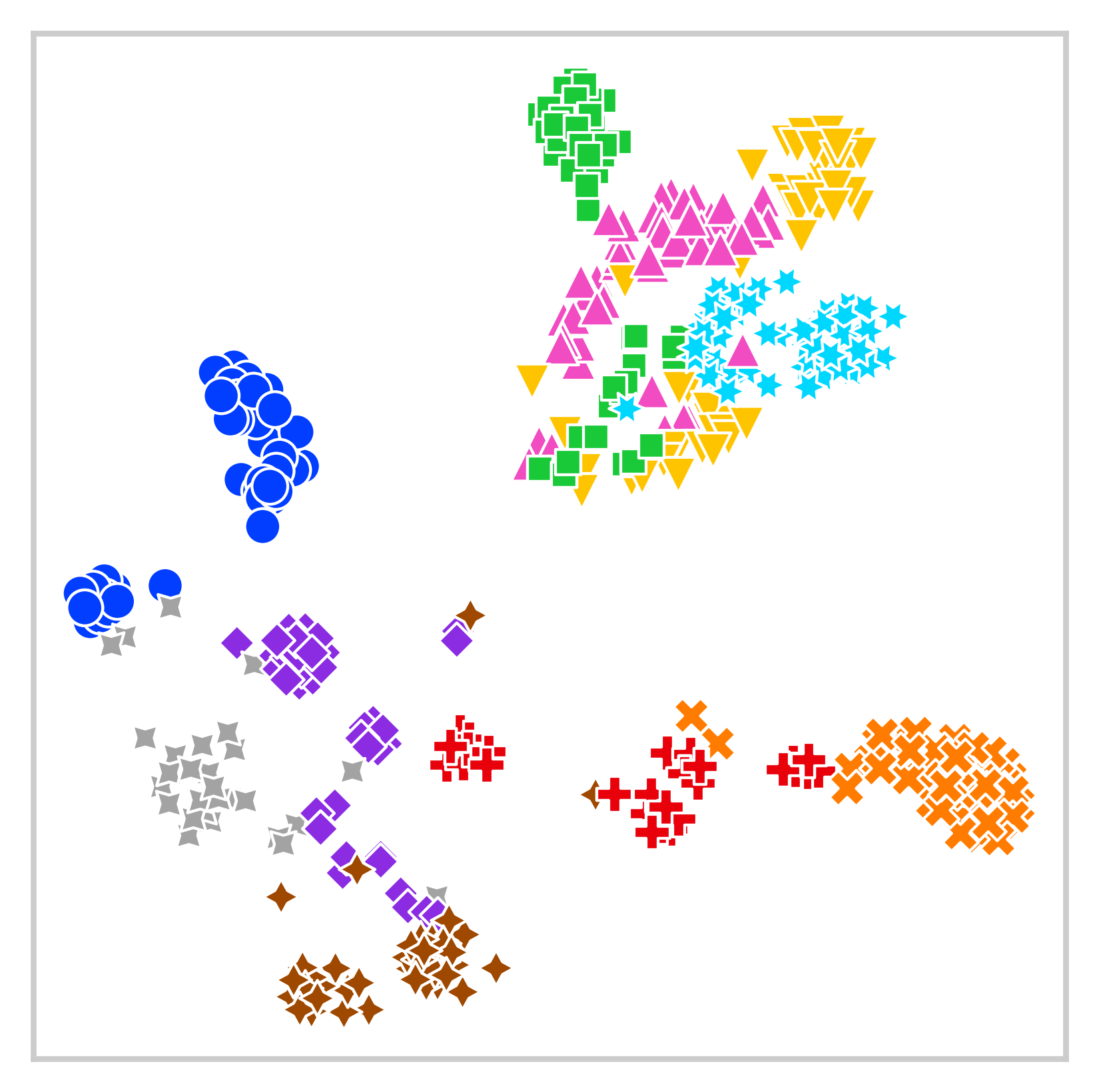} \label{Fig:Track-Embedding:memory}}
        \subfigure[\textit{w/ memory} and \textit{w/ attention} (ours)]{\includegraphics[width=0.24\linewidth]{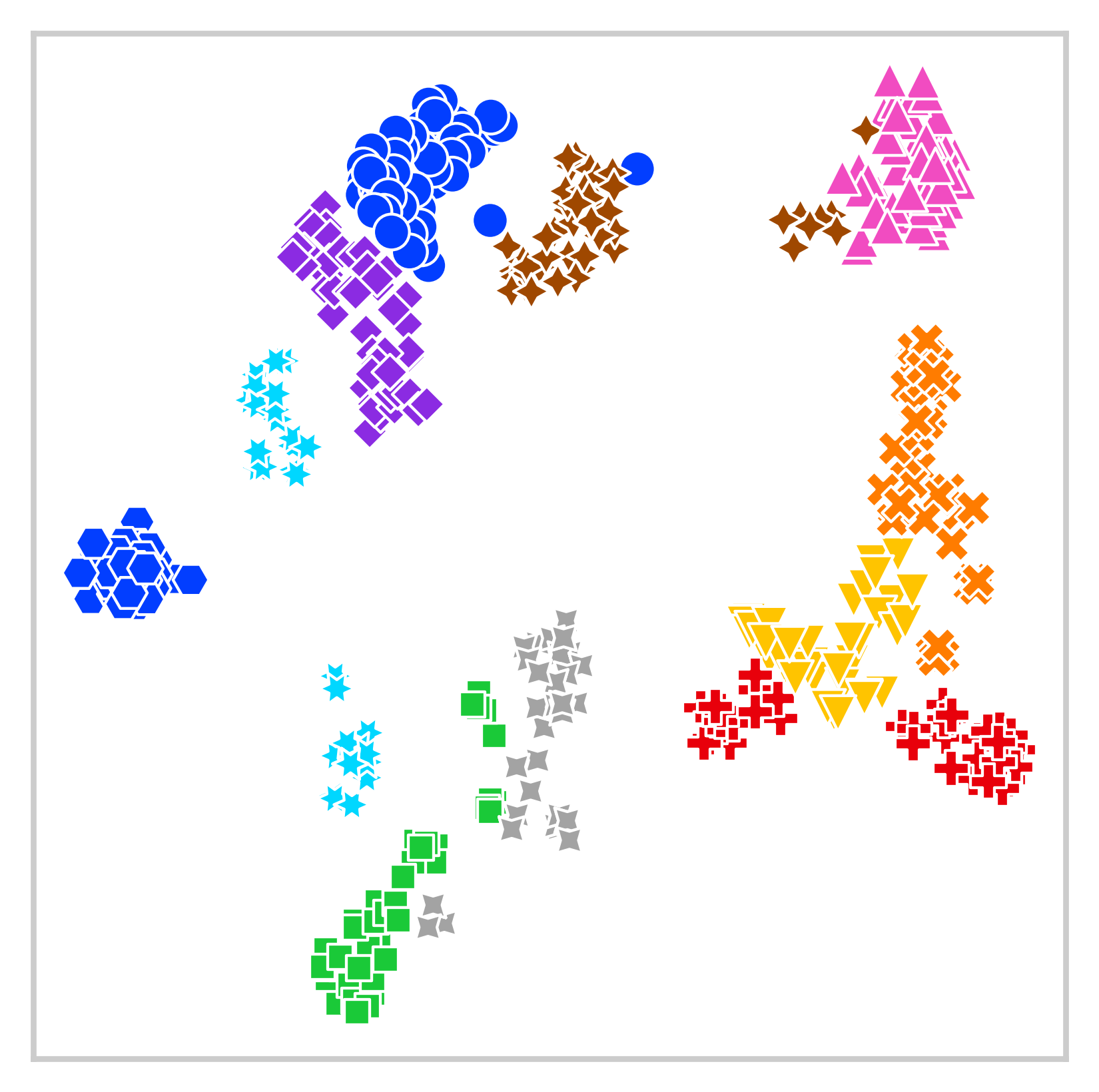} \label{Fig:Track-Embedding:memory+attention}}
	\caption{\textbf{Visualization of Track Embedding $E_{tck}^t$} (the first $50$ frames in sequence dancetrack0063) from different structure designs by using t-Distributed Stochastic Neighbor Embedding (t-SNE). Track embeddings for different tracked targets (IDs) are marked in different colors and shapes. Our design~\ref{Fig:Track-Embedding:memory+attention} helps the model learn a more stable and distinguishable representation for the track embedding. 
    Corresponding tracking performance is shown in Table~\ref{Table:Temporal-Interaction}.
    }
    \label{Fig:Track-Embedding}
\end{figure*}

\noindent \textbf{Long-Term Memory.}
We propose a long-term memory in Section~\ref{Section:Long-Term-Memory} to utilize longer temporal information and further inject it into subsequent track embedding to augment the object feature.

We explore the impact of long-term memory $M_{tck}^t$ and show the experimental results in Table~\ref{Table:Temporal-Interaction}.
For a more comprehensive comparison, we also experiment with another track embedding generation structure that removes the memory-attention layer by passing the temporal aggregation $\hat{O}_{tck}^t$ directly to an FFN network.

Our experimental results show that utilizing long-term memory produces a better association performance, with $+0.8\%$ and $+4.3\%$ AssA for \textit{w/o} and \textit{w/ memory-attention layer}.
The injection of long-term memory significantly stabilizes and augments the identity information of each track embedding, as visualized in Figure~\ref{Fig:Track-Embedding:memory} and~\ref{Fig:Track-Embedding:memory+attention}.

\noindent \textbf{Memory Attention.}
In Table~\ref{Table:Temporal-Interaction}, we also ablate memory-attention layer.
It shows that by using memory-attention layer, our MeMOTR achieves much better performance ($63.9$ \textit{vs.} $62.5$ HOTA), especially improving AssA by $2.1\%$.
This attention layer establishes interactions between different trajectories that help the track embedding learns discriminative features.
However, without long-term memory $M_{tck}^t$, memory-attention layer produces worse performance ($-1.4\%$ AssA and $-1.7\%$ IDF1). We explain that the track embedding without memory augmentation is unstable. Therefore, interacting with such unreliable information can be counterproductive, as visualized in Figure~\ref{Fig:Track-Embedding:attention}.

\noindent \textbf{Memory Update Rate.}
Here, we explored the impact of long-term memory update rate $\lambda$ on the tracking performance in Equation~\ref{Eq:Memory-Update}. As shown in Table~\ref{Table:Long-Memory-Lambda}, when progressively increasing the $\lambda_L$ from $0.005$ to $0.04$, our model achieves the highest HOTA score at $\lambda = 0.01$ while DetA score is decreasing slightly.
We suggest the update rate $\lambda$ may be a hyperparameter that needs to be chosen according to different datasets. For example, scenarios with plenty of target non-rigid deformation may need a higher memory update rate to adapt to the rapidly changing features.

\subsection{Limitations}

Although our MeMOTR brings a significant improvement in association performance, detection performance is still a drawback, especially in crowded scenarios (like MOT17~\cite{MOT16}). During experiments, we observed that sometimes newborn objects are suppressed by tracked targets in the self-attention structure, which leads to reduced detection performance. Therefore, resolving this conflict is a crucial challenge for the joint tracking paradigm. 
This may help improve the detection capabilities, which can boost the overall tracking performance of the model, as studied in~\cite{MOTRv2}.
In addition, in pedestrian tracking, existing datasets are still limited in size and diversity. We suggest that training with other simulation datasets (like MOTSynth~\cite{MOTSync}) may alleviate the overfitting problem of our model and achieve better tracking performance.

\section{Conclusion}

We have proposed MeMOTR, an end-to-end long-term memory-augmented Transformer for multi-object tracking. 
Our method builds a stable long-term memory for each tracked object and exploits this memory to augment the representation of track embedding, thus improving its association performance. Furthermore, by leveraging a memory-attention layer, our model makes different targets more distinguishable. As a result, our approach achieves the state-of-the-art performance on MOT benchmarks, especially in scenes with irregular motion patterns. 
Extensive ablation experiments and visualizations demonstrate the effectiveness of our components.
We hope that future work will pay more attention to the use of long-term temporal information for object tracking.

\noindent \textbf{Acknowledgements.}
This work is supported by National Key R$\&$D Program of China (No. 2022ZD0160900), National Natural Science Foundation of China (No. 62076119, No. 61921006), Fundamental Research Funds for the Central Universities (No. 020214380091, No. 020214380099), and Collaborative Innovation Center of Novel Software Technology and Industrialization. Besides, Ruopeng Gao would like to thank Muyan Yang for her social support.

\appendix

\begin{table*}[t]
  \begin{center}
    \begin{tabular}{l|ccccc|ccccc}
      \toprule[2pt]
      \multirow{2}{*}{\# Row} & \multicolumn{5}{c|}{val set} & \multicolumn{5}{c}{test set} \\
       & HOTA    & DetA    & AssA & MOTA & IDF1 & HOTA    & DetA    & AssA & MOTA & IDF1 \\
      \midrule[1pt]
      \rowcolor{pink!40} 1. MOTR (baseline) & 51.7 & 69.4 & 38.7 & 75.6 & 49.7 & 54.2 & 73.5 & 40.2 & 79.7 & 51.5 \\
      \midrule[1pt]
      \rowcolor{pink!40} 2. \#1 + \textit{memory-augment} & 56.5 & 70.4 & 45.5 & 78.4 & 58.8 & 62.5 & 77.0 & 50.9 & 85.1 & 63.5 \\
      \rowcolor{pink!40} 3. \#2 + ${\mathcal{L}_{d}},{\mathcal{L}_{j}} = 1,5$ & 61.0 & 71.2 & 52.5 & 79.2 & 64.1 & 63.4 & 77.0 & 52.3 & 85.4 & 65.5 \\
      \rowcolor{pink!40} 4. \#3 + Anchor & 61.1 & 73.0 & 51.3 & 81.3 & 63.8 & 64.6 & 78.4 & 53.4 & 87.6 & 67.3 \\
      \rowcolor{cyan!20} 5. \#2 + DAB-D-DETR & 62.1 & 74.3 & 52.2 & 83.1 & 65.6 & 65.9 & 78.8 & 55.2 & 87.9 & 68.9 \\
      \rowcolor{cyan!20} \textbf{6.} \#5 + ${\mathcal{L}_{d}},{\mathcal{L}_{j}} = 1,5$ & \bf 63.9 & \bf 74.6 & \bf 55.0 & \bf 83.4 & \bf 67.1 & \bf 68.5 & \bf 80.5 & \bf 58.4 & \bf 89.8 & \bf 71.2 \\
      \rowcolor{cyan!20} 7. \#5 + ${\mathcal{L}_{d}},{\mathcal{L}_{j}} = 2,4$ & 63.2 & 73.8 & 54.3 & 81.9 & 65.8 & 66.2 & 80.2 & 54.8 & 89.5 & 68.7 \\
      \rowcolor{gray!10} \textcolor{gray}{8. \#6 + YOLOX~[11]} & \textcolor{gray}{66.8} & \textcolor{gray}{78.7} & \textcolor{gray}{57.0} & \textcolor{gray}{88.1} & \textcolor{gray}{70.5} & \textcolor{gray}{70.0} & \textcolor{gray}{81.8} & \textcolor{gray}{60.1} & \textcolor{gray}{90.3} & \textcolor{gray}{72.5} \\
      \bottomrule[2pt]
    \end{tabular}
  \end{center}
  \caption{Supplemental comparison on DanceTrack~\cite{DanceTrack}. Best viewed in color. The same base color results represent using the same DETR framework (D-DETR~\cite{DeformableDETR} or DAB-D-DETR~\cite{DAB-DETR}). $\mathcal{L}_d$ and $\mathcal{L}_j$ are the numbers of detection and joint decoder layers in Figure~\ref{Fig:Overview}, respectively. It should be noted that except for the baseline (\#1), training augmentations (track query erasing and false positive inserting in MOTR~\cite{MOTR}) are removed from other experiments.}
  \label{Table:Boosting}
\end{table*}

\section{Boosting Tracking Performance}

For the tracking-by-detection paradigm, with the development of the Object Detection task, they upgraded the detector used in MOT from Faster R-CNN~\cite{Faster-RCNN} to YOLOX~\cite{YOLOX} and obtained impressive detection performance. Although Deformable-DETR~\cite{DeformableDETR} has competitive detection performance, it still lags behind some popular detectors such as YOLOX~\cite{YOLOX}. This will impair the final tracking performance.

Recently, unlike the original Deformable-DETR~\cite{DeformableDETR}, some methods~\cite{MOTRv2} generate the position embeddings from the learnable anchors. On the one hand, this design will improve the model's detection performance, as discussed in many object detection studies~\cite{DAB-DETR}. On the other hand, the anchor-based position-prior is quite effective for tracking due to frame continuity. 

Therefore, as discussed in Section~\ref{Section:Details}, we built our MeMOTR upon DAB-Deformable-DETR~\cite{DAB-DETR} instead of Deformable-DETR~\cite{DeformableDETR}. We believe that better detection performance of DAB-Deformable-DETR will lead to better tracking performance, as shown in Table~\ref{Table:Boosting} (\#2 \textit{vs.} \#5). We discuss that DAB-Deformable-DETR can be applied in future works as a technology development (like from Faster R-CNN~\cite{Faster-RCNN} to YOLOX~\cite{YOLOX} in the tracking-by-detection paradigm). For a fair comparison with previous transformer-based methods~\cite{TrackFormer, MOTR}, we also provide the results of MeMOTR based on the standard Deformable-DETR in Table~\ref{Table:DanceTrack-SOTA} and~\ref{Table:Boosting} (\#2 and \#3). This indicates our method still has impressive performance without DAB-Deformable-DETR. As done in MOTRv2~\cite{MOTRv2}, we further add the anchor-based position generation process to the standard Deformable-DETR in our method, thus slightly improving the tracking performance (Table~\ref{Table:Boosting} \#3 \textit{vs.} \#4).

Moreover, we also add the YOLOX~\cite{YOLOX} proposal to our model following MOTRv2~\cite{MOTRv2}. As they concluded, this significantly improves the detection and tracking performance simultaneously (Table~\ref{Table:Boosting} \#8). Since the proposals are generated from a frozen CNN-based model, it makes the whole model a non-fully-end-to-end method. For this reason, we list MOTRv2~\cite{MOTRv2} in Table~\ref{Table:MOT17-SOTA} as a new hybrid architecture.

In summary, we provide the cumulative improvements over MOTR~\cite{MOTR} on the val and test set of DanceTrack~\cite{DanceTrack}, as shown in Table~\ref{Table:Boosting}. This further verifies the effectiveness of our various components and gives a more intuitive comparison.

\section{Comparison on Difficult Sequences}

In order to further certify the improvement of our method on target association challenge, we list experimental metrics on some challenging sequences.
We selected eight sequences with the lowest AssA metric of MOTR~\cite{MOTR} on the DanceTrack~\cite{DanceTrack} validation set. 
As shown in Table~\ref{Table:HardSeq}, the association results on these complex sequences are unsatisfactory ($23.6$ average AssA), although the detection performance is passable ($65.8$ average DetA). Our method substantially improves the performance of object association ($35.5$ \textit{vs.} $23.6$ AssA) while slightly improving detection performance ($70.9$ \textit{vs.} $65.8$ DetA).
However, compared to the overall association performance ($58.4$ AssA of our method), there is still a significant deficiency in the results of these challenging sequences.
Therefore, we suggest that improving the object association performance of multi-object tracking is still an unsolved problem that should not be ignored.

\begin{table}[t]
  \begin{center}
  \setlength{\tabcolsep}{0.8mm}{
      \begin{tabular}{c|ccc|ccc}
      \toprule[2pt]
       & \multicolumn{3}{c|}{MOTR~\cite{MOTR}} & \multicolumn{3}{c}{MeMOTR \textit{(ours)}}\\
      Sequence & HOTA & DetA & AssA & HOTA & DetA & AssA \\
      \midrule[1pt]
      dancetrack0041 & 30.6 & 50.9 & 18.8 & 32.9 & 49.4 & 22.4 \\
      dancetrack0081 & 35.7 & 63.9 & 20.0 & 46.0 & 69.6 & 30.4 \\
      dancetrack0063 & 30.5 & 44.9 & 20.7 & 43.2 & 57.4 & 32.6 \\
      dancetrack0019 & 40.8 & 79.3 & 21.0 & 49.7 & 84.2 & 29.3 \\
      dancetrack0014 & 43.7 & 79.4 & 24.2 & 47.6 & 80.0 & 28.4 \\
      dancetrack0004 & 44.1 & 77.0 & 25.3 & 66.2 & 82.7 & 53.0 \\
      dancetrack0034 & 42.3 & 65.7 & 27.3 & 56.9 & 73.2 & 44.3 \\
      dancetrack0090 & 45.1 & 65.1 & 31.5 & 55.4 & 70.5 & 43.7 \\
      \midrule[1pt]
      \textit{average} & 39.1 & 65.8 & 23.6 & \bf \bf \bf 49.7 & \bf \bf 70.9 & \bf 35.5 \\
      \bottomrule[2pt]   
      \end{tabular}
  }
  \end{center}
  \caption{Comparison on difficult sequences on DanceTrack~\cite{DanceTrack} validation set. The results of MOTR~\cite{MOTR} are obtained from the checkpoint provided by the official repo.}
  \label{Table:HardSeq}
\end{table}

\begin{figure*}[t]
	\centering
        \subfigure[\textit{w/o memory} and \textit{w/o attention}]{\includegraphics[width=0.243\linewidth]{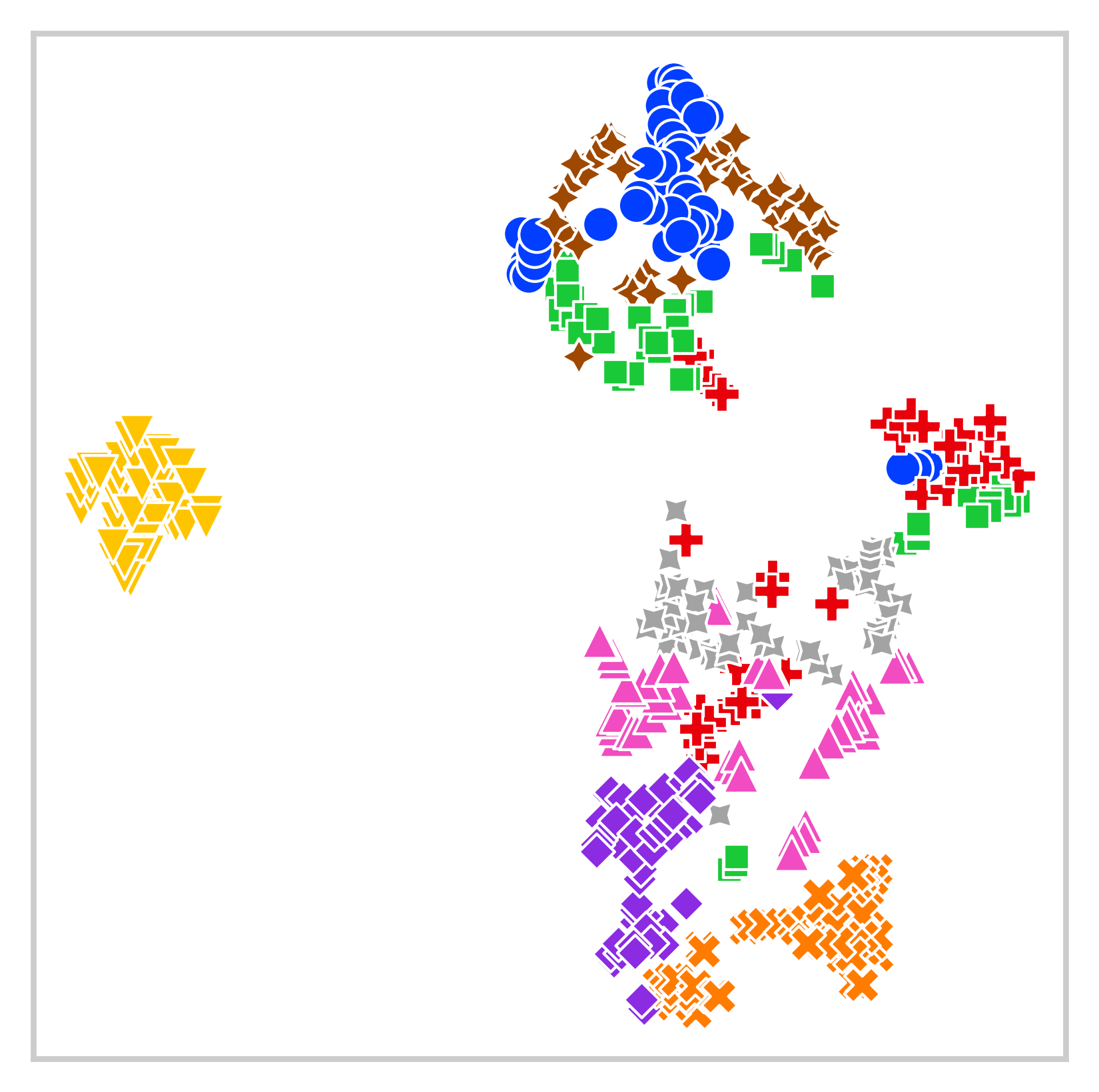}\label{0025:begin}} 
        \subfigure[\textit{w/o memory} and \textit{w/ attention}]{\includegraphics[width=0.243\linewidth]{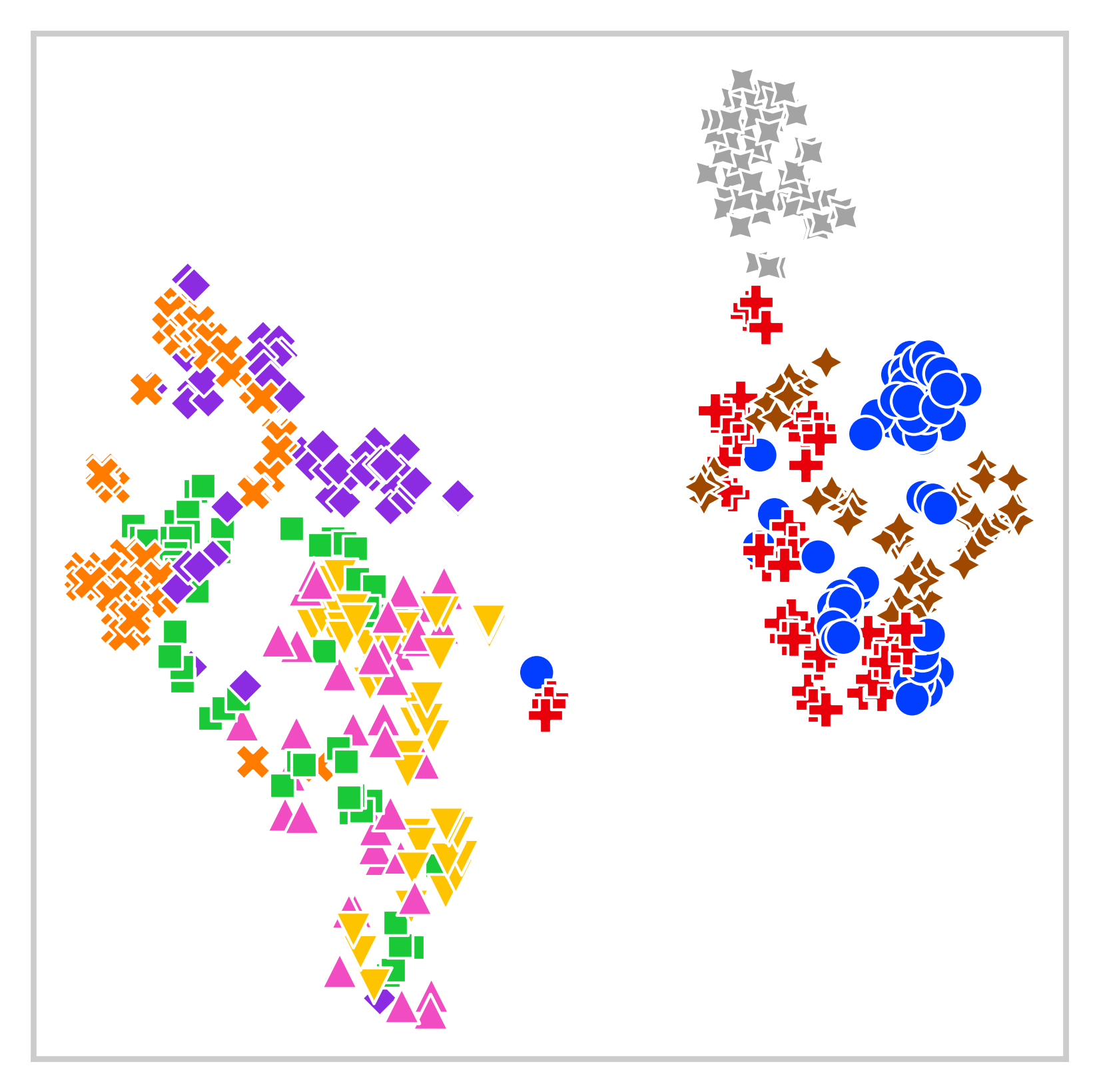}}
        \subfigure[\textit{w/ memory} and \textit{w/o attention}]{\includegraphics[width=0.243\linewidth]{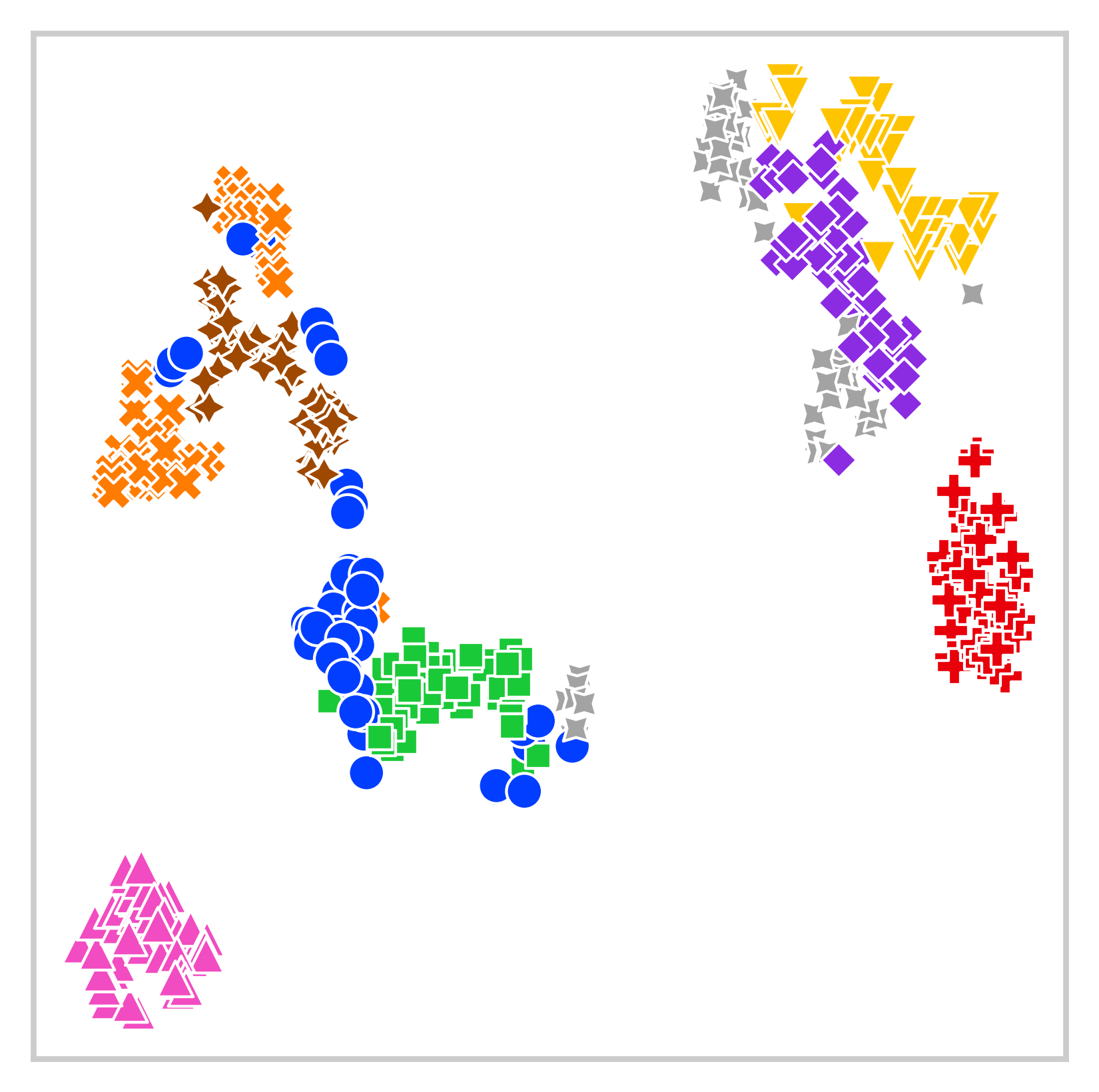}}
        \subfigure[\textit{w/ memory} and \textit{w/ attention} (ours)]{\includegraphics[width=0.243\linewidth]{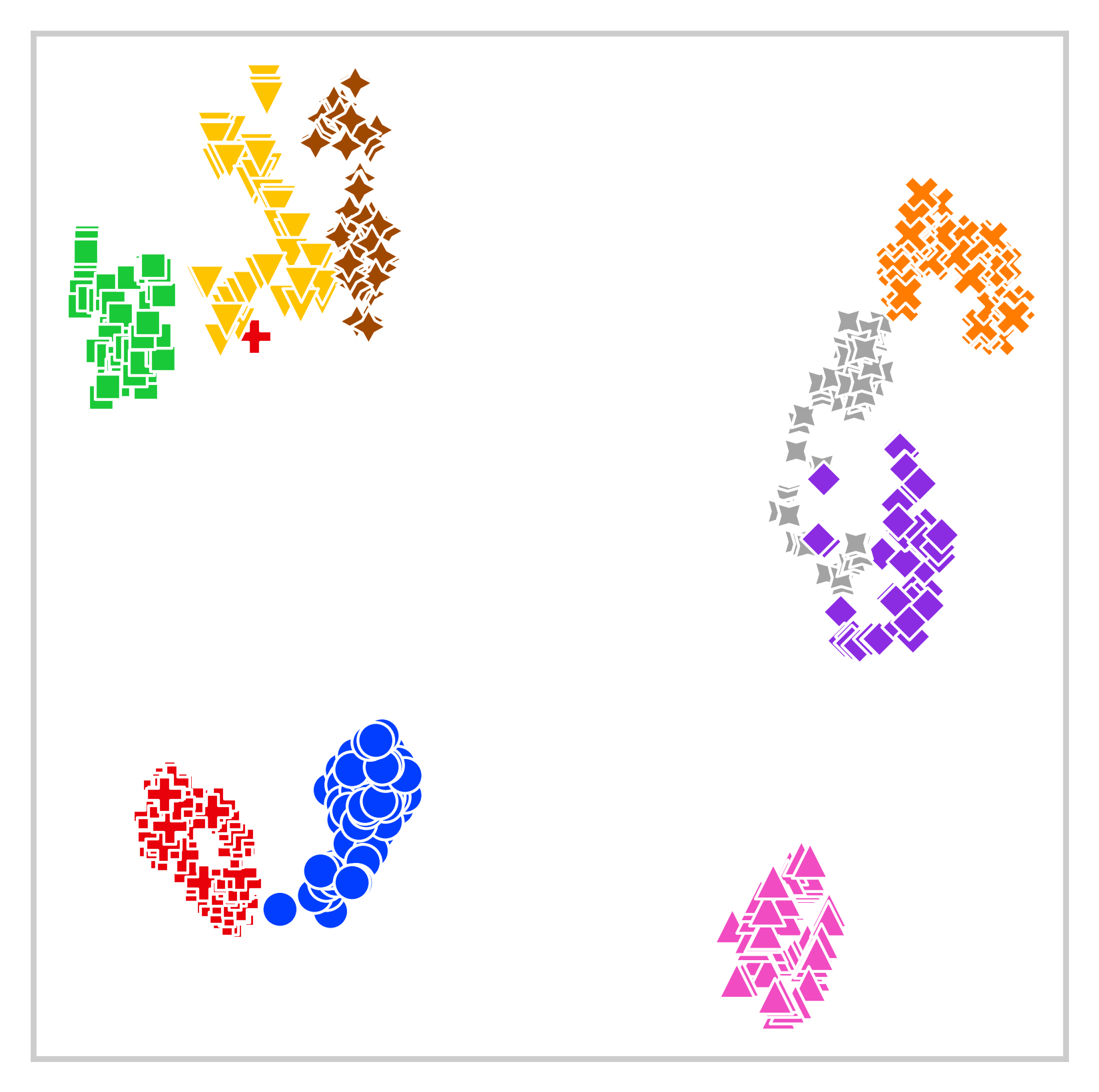}\label{0025:end}} \\
        \subfigure[\textit{w/o memory} and \textit{w/o attention}]{\includegraphics[width=0.243\linewidth]{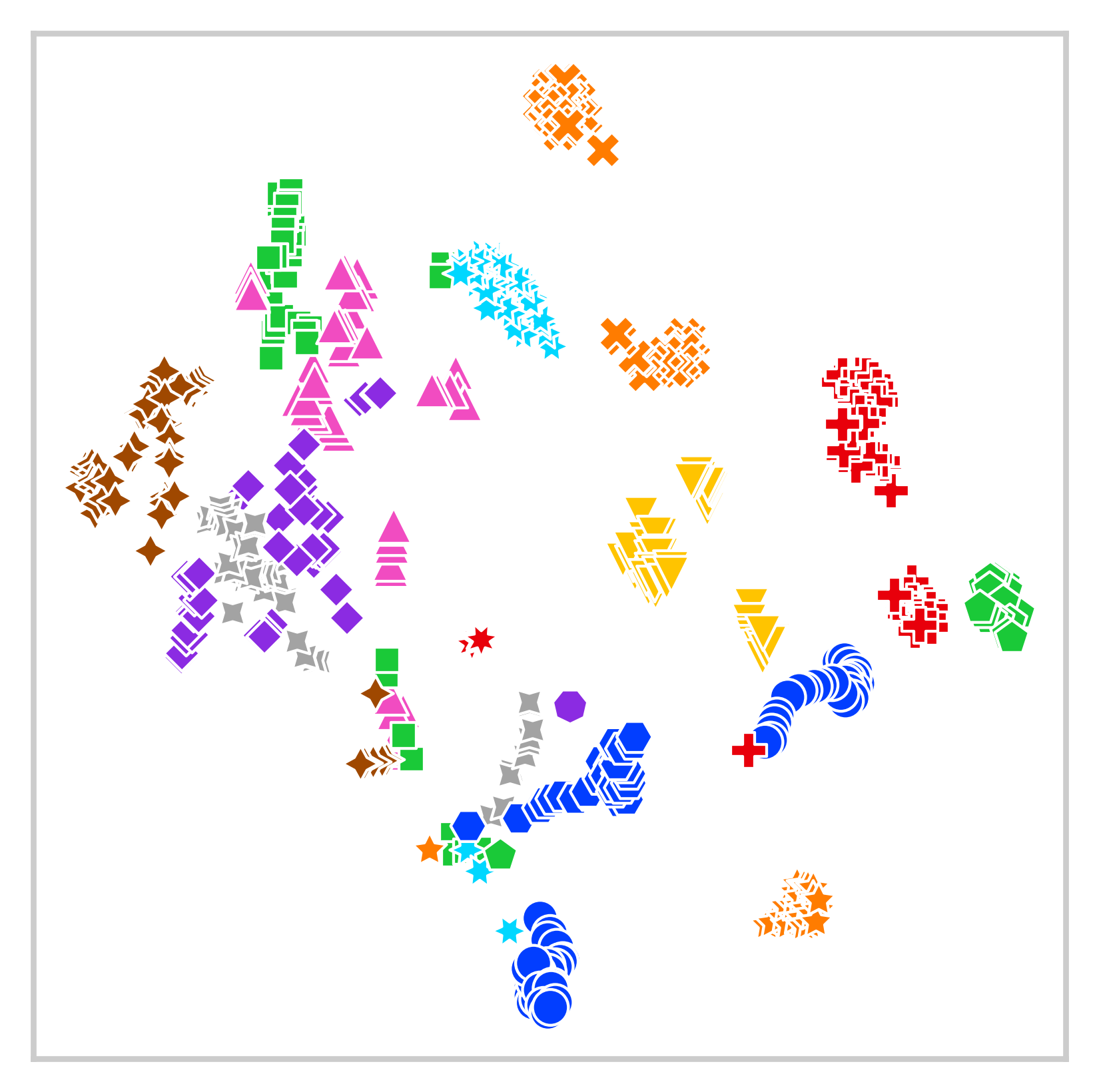}
        \label{0034:begin}} 
        \subfigure[\textit{w/o memory} and \textit{w/ attention}]{\includegraphics[width=0.243\linewidth]{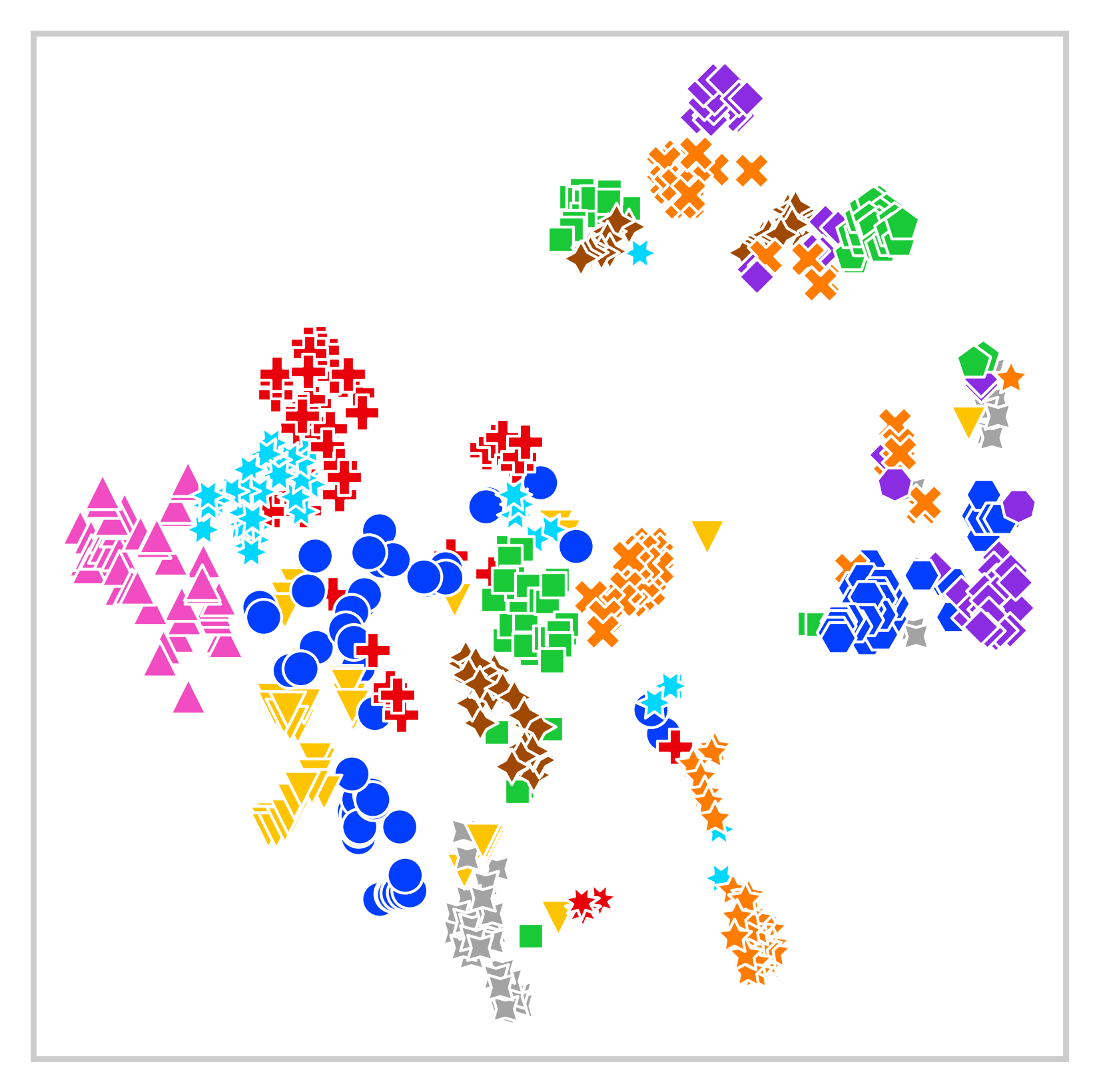}}
        \subfigure[\textit{w/ memory} and \textit{w/o attention}]{\includegraphics[width=0.243\linewidth]{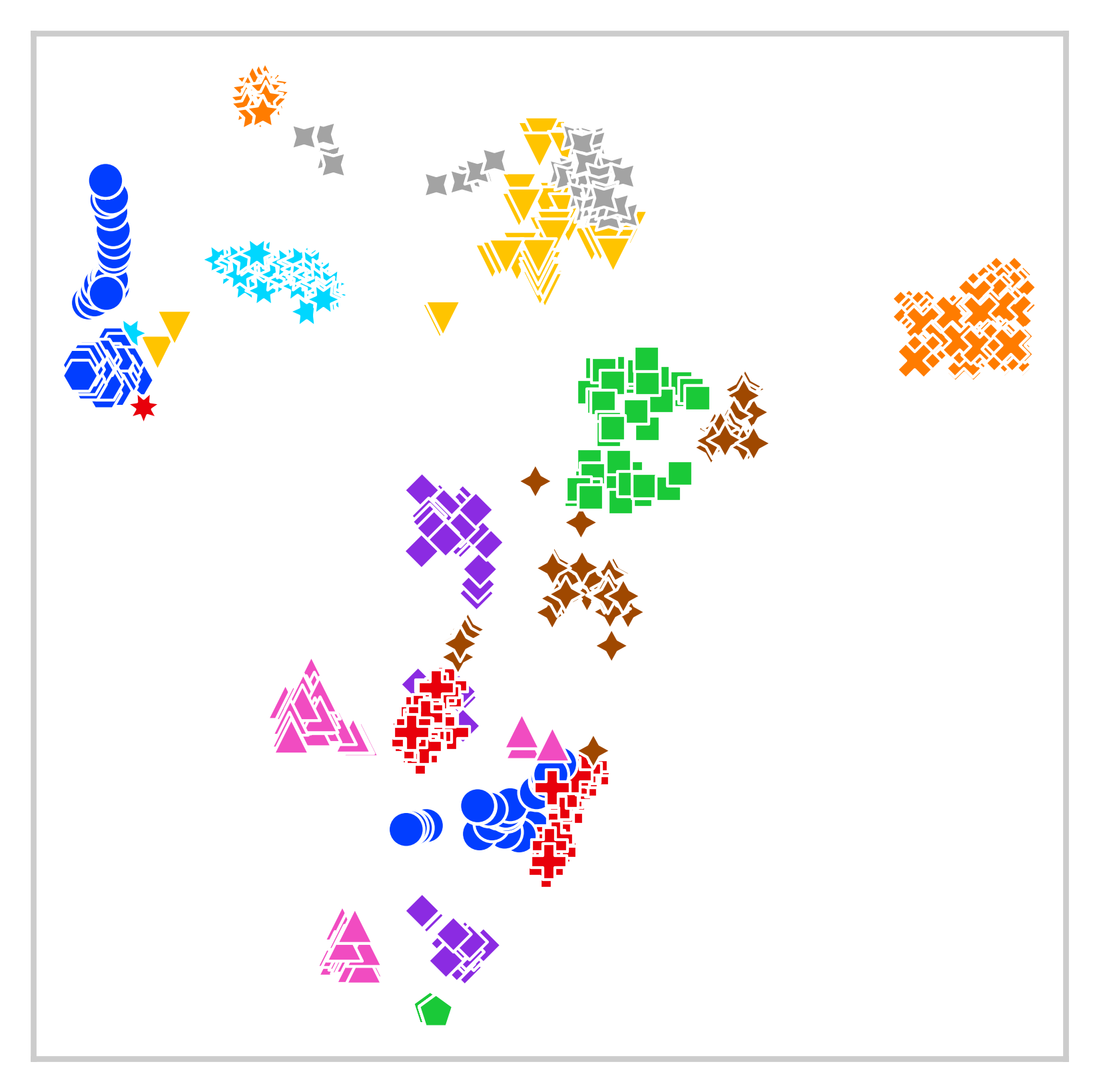}}
        \subfigure[\textit{w/ memory} and \textit{w/ attention} (ours)]{\includegraphics[width=0.243\linewidth]{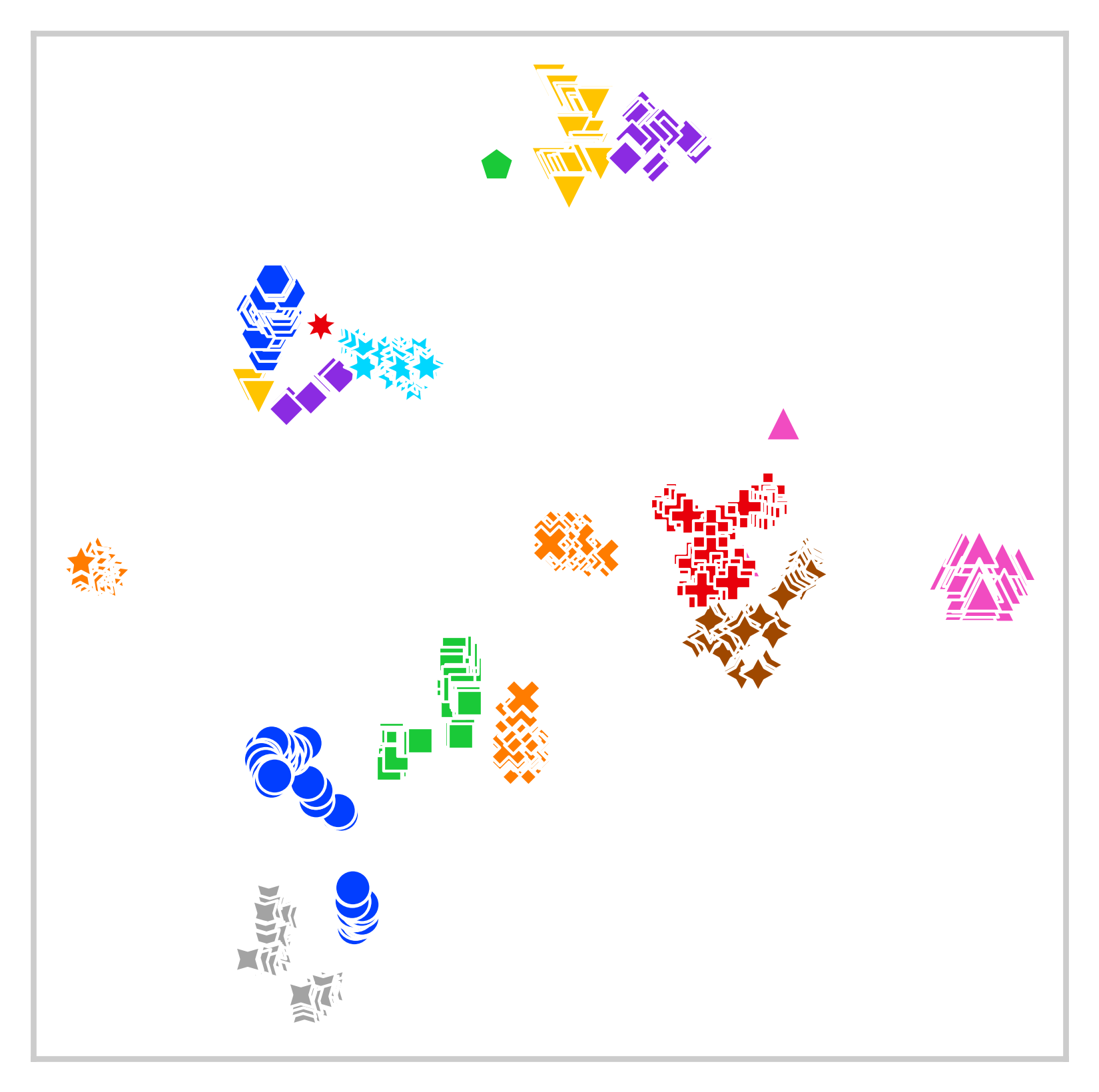}\label{0034:end}}
	\caption{Visualizing track embedding $E_{tck}^t$ from the first 50 frames of \textit{dancetrack0025} (upper) and \textit{dancetrack0034} sequences (lower). Track embeddings for different tracked targets (IDs) are marked in different colors and shapes. The visualizations of our method are shown in Figure~\ref{0025:end} and~\ref{0034:end}.}
    \label{Fig:Embed}
\end{figure*}

\section{More Visualizations}

In this section, we supply additional visualization results. Same as Figure~\ref{Fig:Track-Embedding} in our main paper, we utilize t-Distributed Stochastic Neighbor Embedding (t-SNE) to visualize track embeddings. More visualizing results are provided in Figure~\ref{Fig:Embed}, the upper (Figure~\ref{0025:begin} to~\ref{0025:end}) is from \textit{dancetrack0025}, and the lower (Figure~\ref{0034:begin} to~\ref{0034:end}) is from \textit{dancetrack0034} sequence. These results further verify that our \textit{long-term memory} and \textit{memory-attention layer} help learn a more stable and distinguishable representation for the tracked target.

\begin{table}[t] \small
  \begin{center}
  \setlength{\tabcolsep}{3pt}{
    \begin{tabular}{l|ccccc}
      \toprule[2pt]
      Methods & HOTA & DetA & AssA & MOTA & IDF1 \\
      \midrule[1pt]
      \textit{w/o extra data:} \\
      FairMOT~\cite{FairMOT} & 49.3 & 70.2 & 34.7 & 86.4 & 53.5 \\
      QDTrack~\cite{QDTrack} & 60.4 & 77.5 & 47.2 & 90.1 & 62.3 \\
      ByteTrack~\cite{ByteTrack} & 62.1 & 76.5 & 50.5 & \bf 93.4 & 69.1 \\
      OC-SORT~\cite{OC-SORT} & 68.1 & \bf 84.8 & 54.8 & \bf 93.4 & 68.0 \\
      MeMOTR\textsuperscript{*} (ours) & 68.8 & 82.0 & 57.8 & 90.2 & 69.9 \\
      MeMOTR (ours) & \bf 70.0 & 83.1 & \bf 59.1 & 91.5 & \bf 71.4 \\
      \midrule[1pt]
      \textit{\textcolor{gray}{with extra data:}} \\
      \textcolor{gray}{GTR}~\cite{GTR} & \textcolor{gray}{54.5} & \textcolor{gray}{64.8} & \textcolor{gray}{45.9} & \textcolor{gray}{67.9} & \textcolor{gray}{55.8} \\
      \textcolor{gray}{CenterTrack}~\cite{CenterTrack} & \textcolor{gray}{62.7} & \textcolor{gray}{82.1} & \textcolor{gray}{48.0} & \textcolor{gray}{90.8} & \textcolor{gray}{60.0} \\
      \textcolor{gray}{ByteTrack}~\cite{ByteTrack} & \textcolor{gray}{62.8} & \textcolor{gray}{77.1} & \textcolor{gray}{51.2} & \textcolor{gray}{94.1} & \textcolor{gray}{69.8} \\
      \textcolor{gray}{TransTrack}~\cite{TransTrack} & \textcolor{gray}{68.9} & \textcolor{gray}{82.7} & \textcolor{gray}{57.5} & \textcolor{gray}{92.6} & \textcolor{gray}{71.5} \\
      \textcolor{gray}{OC-SORT}~\cite{OC-SORT} & \textcolor{gray}{71.9} & \textcolor{gray}{86.4} & \textcolor{gray}{59.8} & \textcolor{gray}{94.5} & \textcolor{gray}{72.2} \\
      \bottomrule[2pt]
    \end{tabular}
  }
  \end{center}
  \caption{Performance comparison with state-of-the-art methods on the SportsMOT~\cite{sportsmot} test set. The results for existing methods are basically from the SportsMOT paper. We also report the performance of ByteTrack~\cite{ByteTrack} and OC-SORT~\cite{OC-SORT}, which are trained without additional training data by using their official code. MeMOTR\textsuperscript{*} means the result based on standard Deformable-DETR.}
  \label{Table:SportsMOT-SOTA}
\end{table}

\section{SOTA Comparison on SportsMOT}

SportsMOT~\cite{sportsmot} is a recently proposed multi-object tracking dataset focusing on sports scenarios.
We compare our MeMOTR with the state-of-the-art methods on SportsMOT in Table~\ref{Table:SportsMOT-SOTA}.
In the paper of SportsMOT~\cite{sportsmot}, some methods~\cite{OC-SORT, TransTrack, ByteTrack} utilize extra training data, while others~\cite{QDTrack, FairMOT} do not.
We decided not to use additional data for two reasons: On the one hand, we suggest that SportsMOT already has over 25K frames in its official training set, which is enough for our training. On the other hand, shifting static images as a video clip is a helpless yet bad idea, as it deviates from real tracking scenarios, which is not conducive to end-to-end methods as they learn the knowledge directly from the data distribution. By the way, every person in CrowdHuman~\cite{CrowdHuman} is annotated, while only the players are concerned in SportsMOT~\cite{sportsmot}, which will also challenge the joint training strategy.
For a fair comparison, we utilize the official code to report the performance of ByteTrack~\cite{ByteTrack} and OC-SORT~\cite{OC-SORT}, which are only trained on the training set of SportsMOT. As shown in Table~\ref{Table:SportsMOT-SOTA}, our MeMOTR achieves a surprising performance ($70.0$ HOTA and $59.1$ AssA), further confirming the effectiveness of our design.

{\small
\bibliographystyle{ieee_fullname}
\bibliography{egbib}
}

\end{document}